\documentclass[pmlr]{jmlr}

\RequirePackage{graphicx}
\usepackage{booktabs}
\usepackage{longtable}
\usepackage{algpseudocode}
\makeatletter
\def\set@curr@file#1{\def\@curr@file{#1}} 
\makeatother
\usepackage[load-configurations=version-1]{siunitx} 

\usepackage{enumitem}

\theorembodyfont{\upshape}
\theoremheaderfont{\scshape}
\theorempostheader{:}
\theoremsep{\newline}

\jmlrvolume{[VOLUME \# TBD]}
\jmlryear{2026}
\jmlrworkshop{Machine Learning for Healthcare}

\title[Accelerating Reproducible Research in Synthetic EHR Generation]{Accelerating Reproducible Research in Synthetic EHR Generation}

\author{%
  \Name{Jalen Jiang*}, 
  \Name{Chufan Gao*}, 
  \Name{Ethan Rasmussen*},
  \Name{Stephen Z. Xie}, 
  \Name{Jimeng Sun}\\
  \addr University of Illinois Urbana-Champaign\\
  \addr * Denotes Equal Contribution
}

\begin{document}
\maketitle

\begin{abstract}
The generation of high-fidelity synthetic Electronic Health Records (EHR) is crucial for advancing medical research while preserving patient privacy. However, head-to-head comparison of existing generative models is hindered by disjointed codebases, incompatible data loaders, conflicting library dependencies, and inconsistent evaluation protocols. To address these gaps, we introduce a lightweight, end-to-end benchmarking framework for reproducible synthetic EHR evaluation, organized as a unified pipeline spanning data ingestion, standardized model training, and architecture-agnostic evaluation. Our current implementation targets the generation of longitudinal ICD diagnosis codes--the most commonly studied modality in this literature--and is built on the community-maintained PyHealth library. We reimplement and unify strong baselines (MedGAN, CorGAN, PromptEHR, HALO) under full ICD-9 vocabulary granularity, and add a lightweight GPT-2 baseline from the general-purpose sequence-modeling literature. We contribute a rigorous, architecture-agnostic privacy-utility evaluation suite that applies identically to GAN- and transformer-based generators, and report bootstrapped confidence intervals across all metrics. We further analyze the poor long-tailed performance of existing models and discuss the extensibility of our framework beyond diagnosis codes. By lowering the engineering barrier to running, extending, and evaluating under a single pipeline, we introduce a starting point for community-driven reproducibility and benchmarking synthetic  EHR models.

\end{abstract}

\section{Introduction}
\label{sec:intro}

\begin{figure*}[ht]
\centering
\includegraphics[width=\linewidth]{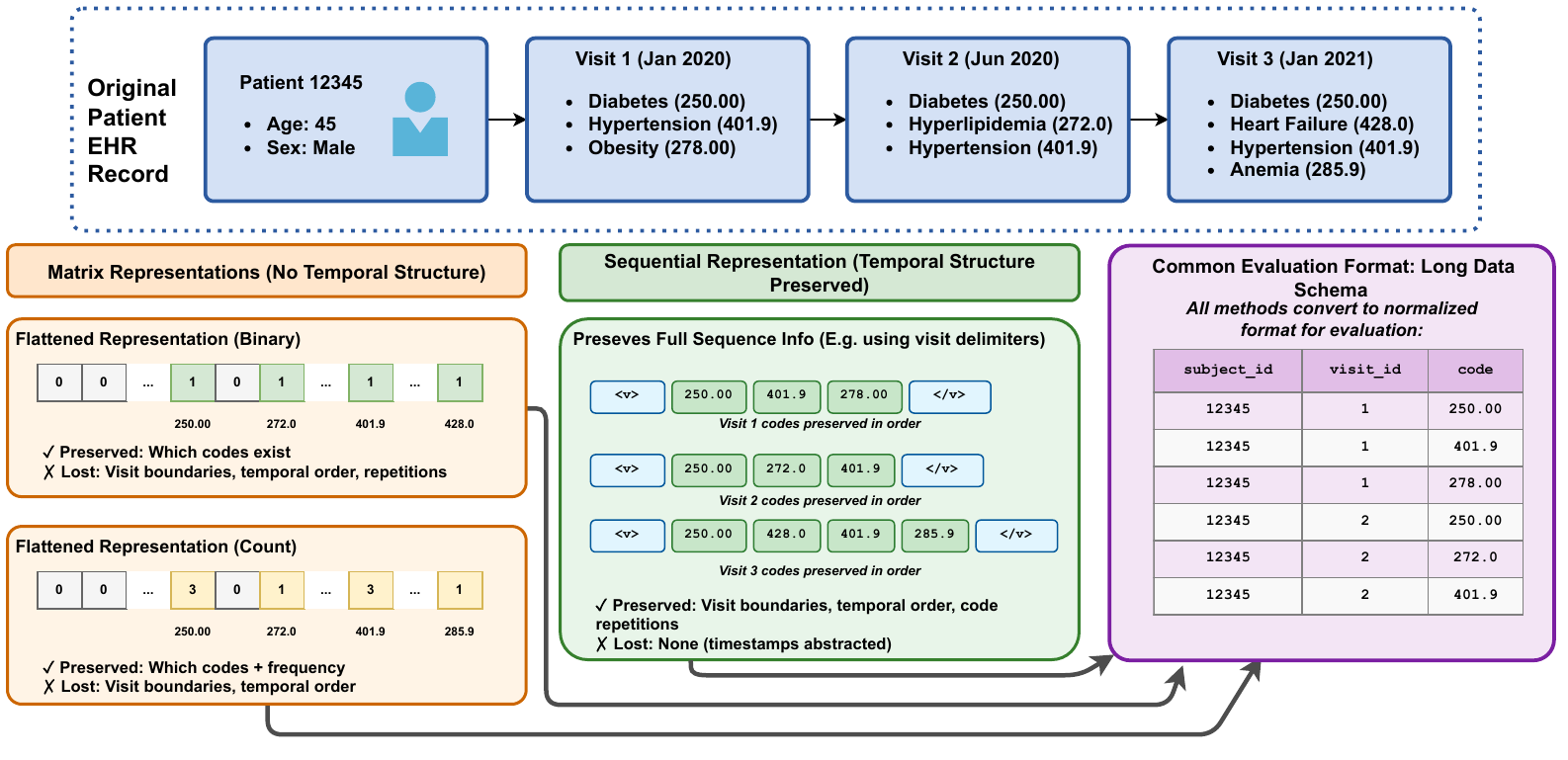}
\caption{\textbf{(Top)} A longitudinal patient trajectory is transformed into two formats for generative modeling. \textbf{(Left)} Flattened approaches remove the temporal dimension. \textbf{(Middle)} In contrast, sequential approaches maintain full sequential information, e.g.\ by using special delimiters for visit boundaries. \textbf{(Right)} To ensure consistent benchmarking, all synthetic outputs are converted into a standardized long-format schema (Subject ID, Visit ID, Code).\label{fig:syn_ehr}}
\vspace{-2em}
\end{figure*}

The proliferation of Electronic Health Record (EHR) systems has created vast repositories of longitudinal patient data with transformative potential for clinical research, from disease progression modeling to treatment optimization using International Classification of Diseases (ICD) codes \citep{nchs9_icd9}. 
Valuable and justified privacy regulations, such as the Health Insurance Portability and Accountability Act (HIPAA) \citep{hipaa1996}, impose severe restrictions on data sharing, which are reasonable yet create a fundamental tension between scientific progress and patient confidentiality \citep{beigi2023simulants, shafquat2023interpretable}. 
Still, this ``data bottleneck" creates a significant obstacle for researchers.
Synthetic data generation offers a principled solution: generate artificial patient records that preserve the statistical properties and correlations of real data while eliminating identifiable information. 

Furthermore, the specialized focus on EHRs has led to an insulation of the field from major advancements in general-purpose tabular data generation. Powerful general synthetic tabular generation models like TabSyn \citep{zhang2023mixed} and TabDiff \citep{shi2024tabdiff} cannot be easily adapted or benchmarked for the unique, complex structure of EHR data. 

Despite significant machine learning methodological advances for EHRs specifically--from autoencoder-based GANs \citep{choi2017generating, torfi2020CorGAN} to transformer-based sequential models \citep{theodorou2023synthesize}--the field faces a reproducibility crisis rooted in several systemic challenges.

\paragraph{Reproducibility Challenges of Disparate ML Methods}

Existing benchmarking efforts face practical barriers to reproducibility and extensibility. SynthEHRella \citep{chen2025generating} aggregates implementations of multiple methods (CorGAN, MedGAN, Synthea \citep{walonoski2018synthea}, and more) but inherits challenges common to research prototypes: method-specific dependency requirements, procedural codebases designed for individual experimental runs. 

A very common roadblock was varying library dependencies. For instance, PromptEHR \citep{wang2022promptehrconditionalelectronichealthcare} integrates evaluation via \texttt{sdmetrics 0.6.0}, but subsequent major version updates (1.x series) introduced breaking API changes, creating forward-compatibility friction.
This technical debt undermines reproducibility: implementations remain procedural, hard-coded to specific dataset dimensions, and unmaintained after publication. When researchers port prior work (e.g., CorGAN's PyTorch reimplementation of MedGAN \citep{torfi2020CorGAN}), implementations often prioritize features aligned with immediate research objectives. This process can lead to unintended feature loss across successive adaptations. For instance, MedGAN's count matrix support was not carried forward in the reference PyTorch implementation.

Architectural capabilities can diverge across implementations of the same method. The original MedGAN \citep{choi2017generating} supported both binary and count matrix representations, implemented in TensorFlow. \citep{torfi2020CorGAN} ports MedGAN to PyTorch as a baseline for their CorGAN study, but they implemented only binary mode as sufficient for their experimental focus, thereby reducing the implementation's generality. Similarly, CorGAN's convolutional architecture was originally configured to 1,071 dimensions, matching the 3-digit ICD-9 truncation standard; adapting it to larger vocabularies required architectural modifications or adaptive pooling strategies. 

This lack of ``out-of-the-box" functionality stifles innovation and makes reproducible comparisons between methods nearly impossible.

\paragraph{A Lack of a Unified, Full-Granularity Evaluation Framework}

More fundamentally, benchmarking practices create structural limitations. The field has sometimes used 3-digit ICD-9 truncation as a shortcut to address sparsity and enable comparison with prior work~\citep{chen2025generating}. Yet medical data distributions inherently follow power laws, and truncating away the tail masks exactly the rare, clinically important codes that distinguish patient subtypes. 

Evaluation metrics—such as privacy metrics (e.g. Adversarial Nearest Neighbor) and utility metrics (e.g. ML privacy estimation), and prevalence statistics—are reported inconsistently, and modern standardized frameworks like SDV's sdmetrics \cite{patki2016synthetic, sdmetrics} are not applicable due to the high dimensionality of ICD codes.

We address these challenges by introducing a unified benchmarking framework built on top of PyHealth~\citep{yang2023pyhealth}, a widely-adopted, community-maintained healthcare ML library with object-oriented design, unit tests, and continuous integration. Concretely, we define our \textit{framework} as an end-to-end pipeline (\textbf{Data $\rightarrow$ Standardized Model $\rightarrow$ Standardized Training $\rightarrow$ Unified Evaluation}) rather than as any single artifact such as a data schema or metric suite. The re-implemented model code, the standardized training recipes, and the architecture-agnostic evaluation code are intrinsically linked: apples-to-apples evaluation is only possible once disparate baselines consume the same inputs and produce outputs in the same long-format schema. We deliberately adopt a visit-level long-format representation that is consistent with existing community conventions (e.g., MEDS~\citep{meds2024}) rather than proposing a new data standard; the novelty of our work lies in the unification of disparate generative models under a single rigorous benchmark, not in the invention of a new way to store patient visits.

\paragraph{Scope.} To keep the comparison tractable, the current instantiation of the framework focuses on the generation of \textit{longitudinal ICD diagnosis codes}, which is the modality that all of our evaluated baselines natively target. We do not claim to generate full multimodal EHRs including labs, vitals, medications, or static demographics; demographics are used as \emph{inputs} (e.g., PromptEHR's age/sex conditioning) rather than as synthesis targets, because no common interface exists across all baselines for generating static patient features. We discuss how the pipeline extends to additional modalities in Section~\ref{sec:extensibility}.

\paragraph{On ``reproducibility'' and software aging.} We do not claim that prior baselines were irreproducible at the time of their release, nor that our framework will permanently withstand technical debt. Our critique is the objective observation that the \emph{current} state of disjointed repositories---pinned to specific dataset dimensions, incompatible dependency versions, and divergent evaluation conventions---makes head-to-head comparison today nearly impossible. To maximize the lifespan of the framework, we rely on: (i) a single installable package rather than per-model forks, (ii) inheritance from the actively-maintained PyHealth library, with its CI and unit tests, (iii) a single language/toolchain across all models, and (iv) an evaluation suite that does not depend on any model-specific loss or output format. We view this framework as a living substrate that the community should iterate on and eventually replace, and we welcome future replacements that move past ICD-9 and PyHealth when those become the limiting factors.

\paragraph{A Lack of Consideration for Long-tailed Codes} 
\begin{figure}[ht]
\centering
\includegraphics[width=.9\linewidth]{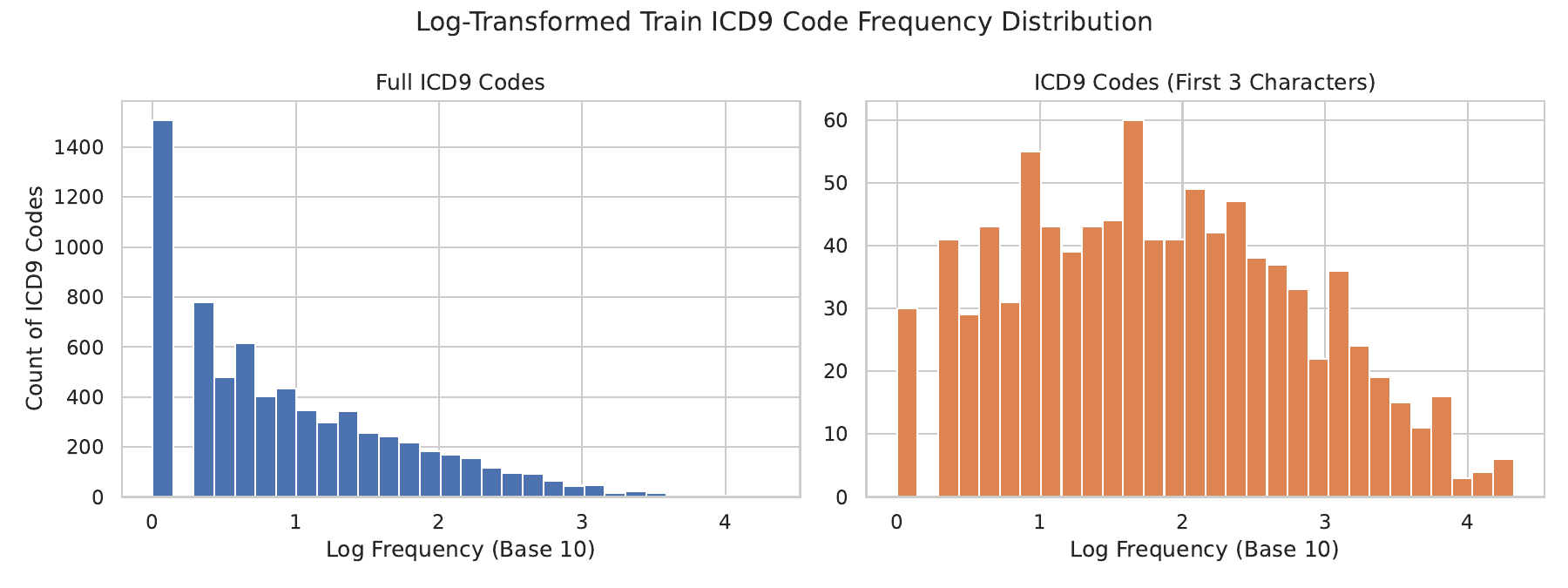}
\caption{Figure that shows that full ICD-9 codes in MIMIC are much more complex and long-tailed. I.e., a minority of common codes dominate the absolute counts of all ICD codes in the patient EHR compared to the simpler first 3 characters. \label{fig:code_distr}}
\vspace{-2em}
\end{figure}
Removing granularity from synthetic codes has become an entrenched norm: researchers commonly reduce ICD-9's 6,955 diagnosis codes to 1,071 three-digit categories, ostensibly to address ``long-tailed" distributions and sparsity (the long-tailed nature is shown in Figure~\ref{fig:code_distr}). Yet medical data is inherently long-tailed; this truncation discards clinically vital information—the distinction between diabetes with renal complications (250.40) and uncomplicated diabetes (250.00) collapses into a single code (250). This practice persists not due to algorithmic necessity, but because benchmarking conventions discourage deviation and limit complete evaluation. Our extended analysis in Appendix~\ref{sec:appendix_marginals} (Figures~\ref{fig:appendix_zipf} and~\ref{fig:appendix_bucket_r2}) confirms that under full-vocabulary evaluation, every generator we test degrades sharply in the rarest deciles of the code distribution---a failure mode that 3-digit-truncated benchmarks systematically hide.

In this work, we present a unified framework for synthetic EHR generation that addresses critical gaps in reproducibility, scalability, and evaluation standardization. Our core contributions are summarized as follows:

\begin{enumerate}
[leftmargin=*, noitemsep]
\item \textbf{Architectural Restoration and Scaling:} We restore and extend the capabilities of canonical baselines to support full-scale ICD-9 vocabularies (6,955 codes), moving beyond the truncated subsets used in prior work. Specifically, we reimplement \textbf{MedGAN}'s~\citep{choi2017generating} count matrix support—omitted in recent PyTorch ports—to enable unbounded integer generation. Furthermore, we extend the \textbf{CorGAN}~\citep{torfi2020CorGAN} architecture from 6 to 8 convolutional layers to natively handle high-dimensional discrete outputs. We also provide best-effort, full-vocabulary implementations of state-of-the-art sequence models, including \textbf{PromptEHR}, \textbf{HALO}~\citep{theodorou2023synthesize}, and a simple GPT2 baseline.

\item \textbf{Standardized Privacy-Utility Benchmarking:} Addressing the fragmentation of assessment metrics in the field, we integrate the \texttt{sdmetrics} library to establish a rigorous evaluation suite. This framework provides consistent, side-by-side comparisons of privacy risks, distributional fidelity, and multivariate correlations across all methods, ensuring fair and reproducible benchmarking.

\item \textbf{Accessible and Extensible Infrastructure:} Built on the popular \texttt{PyHealth} library, our framework democratizes access to complex generative models via a minimal-setup pipeline (single pip install) and interactive Google Colab workflows. The architecture is designed for extensibility, allowing researchers to seamlessly integrate new models and datasets, thereby establishing a solid foundation for future work on ICD-10 migration and richer EHR representations.
\end{enumerate}

\section{Methods}
\subsection{Architectural contributions}

We show the transformation from the raw visit information, the different types of methods we implement, as well as the unified evaluation format in Figure~\ref{fig:syn_ehr}.
To evaluate the impact of temporal granularity on synthetic EHR utility, we analyze two distinct representation paradigms: static matrix-based formulations and dynamic sequential tokenization (see Figure~\ref{fig:syn_ehr}).

Traditional generative models, such as CorGAN and MedGAN, utilize a flattened representation that compresses the longitudinal dimension of patient history into static vectors. As illustrated in Figure~\ref{fig:syn_ehr}, these methods map patient trajectories to fixed-dimensional spaces using either binary indicators (CorGAN) or frequency counts (MedGAN). While this approach effectively preserves global code prevalence—capturing which diagnoses exist within a patient's history--it inherently discards critical temporal structure, thereby limiting their utility for longitudinal clinical analysis.

In contrast, sequential approaches (e.g., PromptEHR) treat the electronic health record as a temporal sequence of discrete tokens and preserve the full temporal fidelity of the record, including the specific ordering of visits and the recurrence of codes across distinct time points.

To ensure fair benchmarking across disparate generative architectures, we standardize all synthetic outputs into a unified long-format schema prior to evaluation. Regardless of whether the source model produces static matrices or dynamic sequences, all data is converted into a normalized relational structure consisting of \texttt{(subject\_id, visit\_id, code)} triplets. This schema is intentionally consistent with established community data formats such as MEDS~\citep{meds2024}; we are not proposing a new data standard, only adopting a standard one so that all benchmarked models share a single, compatible output representation. This post-processing step ensures that metrics capture the structural integrity of the data. For matrix-based models, one visit ID is assigned per patient, highlighting their inability to recover distinct visit clusters. We note that our use of ordinal \texttt{visit\_id}s rather than continuous timestamps is a deliberate simplification: none of the benchmarked baselines produces inter-visit time offsets, so we bin visit order as ordinal IDs to keep the benchmark apples-to-apples across generators. Extending the schema with continuous timestamps is straightforward for any future model that outputs them.

Note that in the process of reimplementing popular baselines, we took significant effort to restore and extend capabilities for all models to support full ICD-9 vocabularies (as this was non-trivial for the flattened models).

\subsection{MedGAN}
For \textbf{MedGAN}~\citep{choi2017generating}, we reimplement count matrix support. Count matrix support was present in the original TensorFlow version but omitted in
subsequent PyTorch ports. This enables generation of unbounded positive integer counts per diagnosis code across all $\sim$6,955 ICD-9 codes.
The original MedGAN supported both binary and count modes, enabling modeling of visit frequencies. The PyTorch implementation provided alongside CorGAN restricted the model to binary mode only, and this version was subsequently adopted by SynthEHRella. Both implementations were limited to three-digit ICD-9 codes, conflating clinically distinct conditions (e.g., diabetes without complications versus diabetes with ketoacidosis both reduce to prefix ``250").

Our implementation extends vocabulary support to full ICD-9 resolution and restores the original count mode capability, enabling modeling of healthcare utilization patterns beyond binary presence/absence. MedGAN's linear layer architecture naturally accommodates arbitrary vocabulary sizes, so we resize to support the larger vocabulary.

\paragraph{Training and Generation.}

The model is trained in two stages: autoencoder pretraining using reconstruction loss, followed by adversarial training with a generator employing residual connections and a discriminator using minibatch averaging. Synthetic
records are produced by sampling latent vectors, passing through the trained generator, and decoding via the autoencoder decoder. Binary mode applies sigmoid activation and thresholds at 0.5; count mode applies ReLU activation
and rounds to integers.

\subsection{CorGAN}
For \textbf{CorGAN}~\citep{torfi2020CorGAN}, we extend the convolutional architecture from
6 layers (hardcoded to 1,071 codes) to 8 layers that natively support 6,955 codes.
We implement CorGAN, a convolutional GAN designed to capture inter-code correlations through learned convolutional filters. The original implementation was configured for three-digit ICD-9 codes with \textit{fixed
convolutional architectures}. Extending to larger vocabularies presents a challenge: CNNs cannot arbitrarily resize outputs without architectural redesign, unlike linear layers.

Our implementation introduces an 8-layer CNN encoder-decoder (versus the original 6 layers) to support full ICD-9 resolution. We empirically validated this architecture against an adaptive pooling approach, finding the 8-layer
design significantly outperforms pooling-based extension with improved prevalence matching and near-zero privacy leakage.

\paragraph{Training and Generation.}

The model is trained in two stages: autoencoder pretraining using binary cross-entropy reconstruction loss, followed by adversarial training using Wasserstein GAN with gradient penalty. Synthetic records are produced by sampling
latent vectors, generating synthetic latent codes via the trained generator, decoding to probability vectors via the trained decoder, and thresholding at 0.5 to obtain binary records.

\subsection{PromptEHR}

We reimplement PromptEHR \citep{wang2022promptehrconditionalelectronichealthcare}, a demographic prompt-conditioned sequence-to-sequence model for generating longitudinal synthetic EHR data. The model employs a BART-based architecture that frames EHR generation as a conditional text generation problem, where patient demographics (age and sex) are encoded as learned prompt embeddings to guide the generation of realistic medical event sequences.
\paragraph{Frequency-Guided Generation.}

The reference PromptEHR implementation addresses prevalence matching through a post-generation augmentation step where a 50\% subset of diagnosis codes from the conditioning patient is incorporated into the synthetic output. While effective at matching training prevalence, this approach limits the degree to which synthetic data is model-generated. We adopt an alternative strategy: frequency-guided generation, where model logits are added with empirical code frequencies during sampling:
$\ell_{\text{guided}} = \ell_{\text{model}} + \alpha \log p_{\text{train}}$

This approach is analogous to how GANs use discriminator feedback to match training distributions, but with an explicit and tunable mechanism. The blending parameter $\alpha$ (empirically set to 2.0) controls the strength of frequency guidance, and the mechanism can be disabled for ablation studies. All codes are model-generated rather than augmented from real patients, supporting privacy-preserving applications while maintaining realistic prevalence patterns.

\paragraph{Demographic-Conditioned Trajectory Initialization.}

To improve clinical realism of generated patient trajectories, we introduce demographic-conditioned first code sampling. The first diagnosis code for each patient visit is sampled from empirical distributions stratified by age (9 bins: 0--10, 10--20, \ldots, 80--90) and sex. This ensures synthetic patients begin with age- and sex-appropriate medical presentations.

\paragraph{Training and Generation.}

\textbf{Training.} The model is trained on MIMIC-III to maximize the conditional log-likelihood of observed patient sequences given demographic prompts. We partition the data into 45,520 training patients for model optimization and 1,000 holdout patients reserved for privacy evaluation. Training employs corruption-based data augmentation (mask infilling and token deletion) with AdamW optimization.

\textbf{Generation.} Synthetic patients are generated through auto-regressive sampling conditioned on demographic prompts sampled from the empirical training distribution. Visit structure (number of visits, codes per visit) is sampled from observed distributions to match real patient trajectory characteristics. During decoding, we apply nucleus sampling (top-$p = 0.95$) with frequency-guided logit blending to generate clinically realistic code sequences.

\subsection{HALO}
HALO~\citep{theodorou2023synthesize} was the only publicly available model that worked reliably with minimal modifications, highlighting the rarity of production-ready implementations in the open-source EHR generation domain.Beyond updating hard-coded paths and extending training to 80 epochs, the reference implementation required no substantial changes. HALO models patient records as a sequence of longitudinal visits, where each visit is represented as a multi-hot binary vector of medical codes. The system employs a hierarchical architecture consisting of a visit-level module to capture long-term patient history and a code-level module to manage intra-visit dependencies. This 2-step, hierarchical approach allows the model to generate high-dimensional ICD codes by first factorizing probabilities across visits and then refining through a fine-grained, autoregressive modeling of codes within each individual visit.

\subsection{GPT Baseline} 
To add completeness in the benchmarking. We want to compare the specialized healthcare generative models against general-purpose sequence modeling architectures, we implemented a lightweight version of the GPT-2 model \citep{radford2019language, wolf2019huggingface}. While GPT-2 is traditionally utilized for natural language processing, its autoregressive capabilities make it a strong candidate for modeling the sequential dependencies inherent in longitudinal patient trajectories.

\paragraph{Implementation Details.}
We adapted the architecture using the Hugging Face \texttt{transformers} library. To accommodate the specific vocabulary size and dataset scale of MIMIC-III, we utilized a reduced configuration compared to the standard decoder-only GPT-2 Small: 8 hidden layers, 8 attention heads, and an embedding dimension of 512 (\texttt{n\_embd=512}). We simply extended the tokenizer to include all new ICD codes as new tokens, as well as a custom \texttt{visit\_delim} code to separate out sequential visits.

Data preprocessing involved converting patient clinical histories into textual sequences, where unique ICD-9 codes function as discrete tokens. We trained a custom \texttt{Word Level} tokenizer to handle the medical vocabulary without sub-word fragmentation, ensuring that diagnosis codes remain intact.

\section{Evaluation Methodology}
\label{sec:methodology}

To rigorously assess the quality of the generated synthetic electronic health records (EHRs), we employ a comprehensive suite of metrics spanning two critical axes: \textit{fidelity} (statistical and functional similarity to real data) and \textit{privacy} (resistance to re-identification and membership inference). Given the longitudinal and high-dimensional nature of EHR data---where patients have variable-length histories of visits, each containing sets of multi-hot disease codes---standard tabular metrics are insufficient. We introduce a specialized distance metric for sequential sets and evaluate our model using the following protocols.

\paragraph{Metric selection rationale.}
A core design constraint for this benchmark is that all metrics must be \emph{architecture-agnostic}: they must be computable from the final, discretized patient records produced by any generator, without relying on internal model quantities. This rules out likelihood- or logit-based metrics (which favor autoregressive transformers over GANs), discriminator-loss-based metrics (which only make sense for adversarial generators), and reconstruction losses (which only apply to autoencoder-based methods). We also avoid metrics that require exact nearest-neighbor density estimation (e.g., $\alpha$-precision / $\beta$-recall~\citep{alaa2022faithful}): on full ICD-9 vocabularies with variable-length visit sequences these are computationally intractable, as noted by recent tabular-synthesis work~\citep{shi2024tabdiff,garuti2025diffusion}. We therefore use a scalable discriminator-based fidelity score (Section~\ref{sec:methodology}) as a tractable proxy for joint-distribution matching, together with marginal-level prevalence metrics and distance-based privacy attacks. We do not claim these are the \emph{only} valid metrics; rather, they are the subset of commonly-used metrics that can be consistently applied to every model in the benchmark. Prevalence is included because, while it is a marginal rather than a joint metric, a generator that cannot even reproduce disease base rates is not useful for downstream epidemiological modeling, cohort construction, or disease-surveillance studies, which are among the primary intended uses of synthetic EHR data.

\subsection{Fidelity and Utility Metrics}

We evaluate whether the synthetic data preserves the statistical properties and predictive signal of the real distribution.

\paragraph{Statistical Code Prevalence:} 
We compute the patient-level prevalence for every unique medical code $c$ in the vocabulary. Let $N$ be the number of unique patients. The prevalence probability is $p(c) = \frac{1}{N}\sum_{i=1}^N \mathbb{I}(c \in P_i)$. We compare the prevalence vectors of the real ($\mathbf{p}_{real}$) and synthetic ($\mathbf{p}_{syn}$) populations using the Coefficient of Determination ($R^2$), Pearson correlation coefficient ($\rho$), and Root Mean Squared Error (RMSE). High correlations indicate the generator correctly captures the marginal distributions of diseases.

\paragraph{Discriminator-Based Fidelity Evaluation}

To quantitatively assess the distributional similarity between the real Electronic Health Records (EHR) and the generated synthetic data, we employ a machine learning discriminator metric. This approach evaluates the "indistinguishability" of the synthetic data, positing that high-fidelity synthetic data should be difficult for a classifier to differentiate from real data.

Let $\mathcal{D}_{\text{real}} = \{x_i\}_{i=1}^{N_r}$ represent the set of real patient sequences and $\mathcal{D}_{\text{syn}} = \{\hat{x}_j\}_{j=1}^{N_s}$ represent the set of synthetic patient sequences. We construct a supervised binary classification task, $\mathcal{D}_{\text{disc}}$, by labeling real samples as class $1$ and synthetic samples as class $0$:
$\mathcal{D}_{\text{disc}} = \{(x, 1) \mid x \in \mathcal{D}_{\text{real}}\} \cup \{(\hat{x}, 0) \mid \hat{x} \in \mathcal{D}_{\text{syn}}\}$

To prevent data leakage, we perform a patient-level stratified split, dividing $\mathcal{D}_{\text{disc}}$ into training and testing sets such that disjoint sets of patients are used for training and evaluation.

We train a binary classifier $f_\theta$ (parameterized by $\theta$) to discriminate between the two sources. The model architecture $f_\theta$ mirrors the downstream utility models (e.g., an LSTM for sequential visit data or a Random Forest for flattened features). The objective is to minimize the binary cross-entropy loss.
The fidelity is measured by the classification accuracy ($Acc_{\text{disc}}$) on the held-out test set. 

\textbf{Ideal Fidelity Score ($Acc_{\text{disc}} \approx 0.5$):} If the classifier performs no better than random guessing, the synthetic data is indistinguishable from the real data, indicating that the generative model has successfully captured the underlying distribution.
\textbf{Low Fidelity Score ($Acc_{\text{disc}} \gg 0.5$):} A high classification accuracy implies that the synthetic data contains distinct artifacts or distributional shifts that allow the model to easily identify fake records.

To provide a normalized metric where higher values indicate better quality (indistinguishability), we calculate a Privacy/Fidelity Score bounded between $[0, 1]: \text{Score} = 1 - 2 \cdot |0.5 - Acc_{\text{disc}}|$,
where a score of $1.0$ corresponds to perfect indistinguishability ($Acc_{\text{disc}} = 0.5$).
Note that we used an LSTM for the sequential models and a Random Forest for the flattened baselines. Details are in Appendix~\ref{sec:appendix_discriminator_eval}.

\subsection{Privacy Metrics}

\subsubsection{A Set Difference Distance Metric for Sequential Multi-hot Data}
\label{sec:distance}
Many privacy metrics rely on calculating the distance between patient records. Our data is structured as an \emph{ordered sequence of unordered sets}: the visits within a patient have a well-defined temporal order, but the diagnosis codes inside any single visit are unordered (each visit is a multi-hot set). Euclidean distance on flattened multi-hot vectors ignores the visit-level temporal structure, so we use a sequence-aware set-distance instead. We define a modified Hamming-like distance, $d(P_A, P_B)$, between two patient records $P_A$ and $P_B$.

Let a patient record $P$ consist of a sequence of visits $V = \{v_1, v_2, \dots, v_T\}$, where each visit $v_t$ is a set of active medical codes. The distance is calculated as the sum of the symmetric differences between aligned visits, plus a penalty for length mismatch:
\begin{align*}
d(P_A, P_B) = \delta_{len} + \sum_{k=\min(T_A, T_B)}^{T_{max}} |v_{k}| + 
\sum_{t=1}^{\min(T_A, T_B)} \left( |v_{A,t}| + |v_{B,t}| - 2|v_{A,t} \cap v_{B,t}| \right)
\end{align*}

where $\delta_{len} = \mathbb{I}(T_A \neq T_B)$ is a binary length penalty, and the final term adds the cardinality of extra visits from the longer sequence to the distance. This ensures that both the content of visits and the temporal trajectory are accounted for.

\textbf{Nearest Neighbor Adversarial Accuracy Risk (NNAAR):} 
To ensure the model is not merely memorizing training data, we employ distance-based privacy attacks using the metric defined in Sec.~\ref{sec:distance}.

As used by \cite{theodorou2023synthesize}, NNAAR quantifies the risk that synthetic samples are closer to the training set than expected. We compute the Adversarial Accuracy ($AA$) score, which measures the likelihood that a sample from dataset $A$ is closer to a neighbor in dataset $B$ than to a neighbor in its own dataset $A$:
{\footnotesize
\begin{align*}
AA_{AB} = \frac{1}{2} \left( \frac{1}{N}\sum_{i=1}^N \mathbb{I}(d_{AB}^{(i)} > d_{AA}^{(i)}) + \frac{1}{N}\sum_{i=1}^N \mathbb{I}(d_{BA}^{(i)} > d_{BB}^{(i)}) \right)
\end{align*}
}
The NNAAR score is defined as $AA_{ES} - AA_{TS}$, where $E$ is the held-out evaluation (test) set, $S$ is the synthetic set, and $T$ is the training set. A positive score suggests overfitting (synthetic data is closer to training data than general evaluation data), while a score near zero (or $<0.03$) indicates acceptable privacy risks.

\textbf{Membership Inference Attack (MIA):} 

We simulate a dataset attack to determine if an adversary can distinguish members of the training set from non-members. We construct an attack dataset containing equal numbers of training records (members) and held-out test records (non-members). For each record $x$, we compute its distance to the nearest neighbor in the synthetic dataset: $d_{min}(x, \mathcal{D}_{syn}) = \min_{s \in \mathcal{D}_{syn}} d(x, s)$.

We apply a threshold attack where records with $d_{min}$ lower than the median distance are predicted as members. We report the Accuracy, Precision, Recall, and F1-score of this attack. An accuracy near 0.5 indicates the synthetic data does not leak membership information via distance proximity.

\begin{table*}[ht]
\centering
\caption{Comparative Evaluation of Synthetic Data Generation Methods. Values are reported as $\text{mean}_{\text{std}}$ across 5 bootstrap resamples of the synthetic and evaluation sets. The table categorizes results into Privacy Risk (0.50 is ideal for MIA; NNAAR near 0 is best), Statistical Utility (higher $R^2$ is better), and Fidelity (Discriminator Accuracy close to 0.50 is ideal; higher Discriminative Score is better).}
\label{tab:synthetic_eval}
\resizebox{.9\linewidth}{!}{%
\begin{tabular}{lcccccc}
\toprule
& \multicolumn{3}{c}{\textbf{Privacy Risk Metrics}} & \multicolumn{3}{c}{\textbf{Fidelity / Utility}}  \\
\cmidrule(lr){2-4} \cmidrule(lr){5-7}
\textbf{Method} & NNAAR & \textbf{MIA F1} & \textbf{MIA Acc.} & \textbf{Prev. $R^2$} & \textbf{Prev. RMSE} & \textbf{Disc. Score} \\
& (Target $\approx 0$) & (Target $\approx 0.5$) & (Target $\approx 0.5$) & ($\uparrow$ Better) & ($\downarrow$ Better) & ($\uparrow$ Better) \\
\midrule
\multicolumn{7}{l}{\textit{\textbf{Sequential Models}}} \\
\midrule
HALO          & $0.0060_{0.0034}$  & $0.5000_{0.0158}$ & $0.5140_{0.0118}$ & $0.9545_{0.0104}$ & $0.0021_{0.0001}$ & $0.3266_{0.0076}$ \\
PromptEHR     & $0.0000_{0.0006}$  & $0.5162_{0.0165}$ & $0.5220_{0.0115}$ & $0.8655_{0.0388}$ & $0.0036_{0.0005}$ & $0.3241_{0.0034}$ \\
GPT Baseline  & $0.0085_{0.0057}$  & $0.5047_{0.0077}$ & $0.5290_{0.0040}$ & $0.8729_{0.0327}$ & $0.0035_{0.0003}$ & $0.3550_{0.0049}$ \\
\midrule
\multicolumn{7}{l}{\textit{\textbf{Flat Models}}} \\
\midrule
MedGAN        & $0.0015_{0.0069}$  & $0.4925_{0.0156}$ & $0.4930_{0.0164}$ & $0.9788_{0.0044}$ & $0.0014_{0.0001}$ & $0.0560_{0.0052}$ \\
CorGAN        & $-0.0010_{0.0010}$ & $0.4930_{0.0150}$ & $0.5270_{0.0089}$ & $0.6235_{0.0730}$ & $0.0061_{0.0003}$ & $0.1310_{0.0010}$ \\
\bottomrule
\end{tabular}%
}
\end{table*}

\section{Results}

\paragraph{Models Achieve Imbalanced Fidelity and Distributional Similarity Metrics}
Our evaluation of the Discriminative Score (derived from the discriminator's inability to distinguish real from synthetic data) reveals a distinct dichotomy between the sequential and flat modeling approaches. 

The sequential models (HALO, PromptEHR, and GPT Baseline) demonstrated superior fidelity compared to their flat counterparts in terms of \textit{Discriminative Scores}. The LSTM discriminator struggled more significantly to differentiate these synthetic sequences from real patient trajectories, resulting in higher Discriminative Scores. This suggests that the sequential generators successfully approximated the joint probability distribution of the data, capturing both the co-occurrence of codes within visits and the transition probabilities across time. 

In contrast, the flat models (MedGAN and CorGAN), despite achieving high statistical utility on marginal counts, yielded significantly lower Discriminative Scores. The Random Forest discriminator was able to distinguish synthetic records from real records with high accuracy. This observation points to a common failure mode in GAN-based tabular generation: while the models may perfectly match the \textit{marginal} distributions (e.g., the prevalence of individual diseases), they often fail to capture the complex, high-dimensional \textit{joint} correlations, leaving distinct artifacts that a discriminative classifier can easily exploit. Appendix~\ref{sec:appendix_cooc} (Figure~\ref{fig:appendix_cooc}, Table~\ref{tab:appendix_cooc}) corroborates this directly: on the top-50 ICD-9 codes, the pairwise co-occurrence matrices of HALO and MedGAN are within 0.03 cosine of the real matrix, while CorGAN and PromptEHR's are noticeably weaker ($<0.85$). Appendix~\ref{sec:appendix_structure} further shows that flat models only reproduce codes-per-patient because all codes are pooled into a single pseudo-visit, which trivializes the structural axis of the benchmark.

\paragraph{Models Satisfy Privacy Metrics}
While fidelity metrics varied significantly across model architectures, the privacy risk metrics—specifically Nearest Neighbor Adversarial Accuracy Risk (NNAAR) and Membership Inference Attack (MIA)—showed a consistent trend of robustness across all evaluated methods. 

Regardless of the generator's architecture (sequential or flat), the MIA accuracy hovered near random guessing, and NNAAR scores remained negligible. This indicates that none of the models exhibited significant memorization of the training data. 

Notably, we observed an inverse relationship between statistical utility and fidelity in the flat models. While methods like MedGAN achieved near-perfect code prevalence alignment (Utility), this came at the cost of poor indistinguishability (Fidelity). Conversely, the sequential models struck a more balanced trade-off, sacrificing a small degree of marginal alignment for a significantly more realistic representation of the joint distribution and temporal dynamics, without compromising patient privacy.

\paragraph{GPT Baseline Works Well}
Despite its implementation simplicity relative to all other domain-specific models, the GPT-2 baseline demonstrated remarkable efficacy. As shown in Table \ref{tab:synthetic_eval}, the model achieved a Discriminative Score of \textbf{0.3550}, the highest among all evaluated sequential methods. This indicates that the distributions generated by the GPT-2 baseline were the most challenging for the LSTM discriminator to distinguish from real patient data, suggesting a superior capture of complex temporal dynamics.

In terms of statistical utility, the model remained competitive with an $R^2$ of 0.8729, closely trailing PromptEHR ($0.8655$) and HALO ($0.9545$). Critically, this high fidelity was achieved without compromising privacy; the model maintained a negligible Nearest Neighbor Adversarial Accuracy Risk (NNAAR) and Membership Inference Attack (MIA) F1 score, effectively equivalent to random guessing. These results highlight that standard, well-tuned autoregressive transformers can serve as formidable baselines for EHR synthesis, often outperforming complex, specialized architectures in realism (Fidelity) even if they slightly lag in marginal code prevalence (Utility). The per-decile decomposition in Appendix~\ref{sec:appendix_marginals} (Figure~\ref{fig:appendix_bucket_r2}) shows that the GPT baseline's aggregate $R^2$ is driven almost entirely by head-of-distribution codes---its fit collapses in the rarest 30\% of codes---which explains how a model with only 2{,}244 unique codes emitted (Table~\ref{tab:appendix_coverage}) can still be competitive on an aggregate metric.

\paragraph{Long Tailed Codes}

\begin{figure*}[ht]
\centering
\includegraphics[width=.9\linewidth]{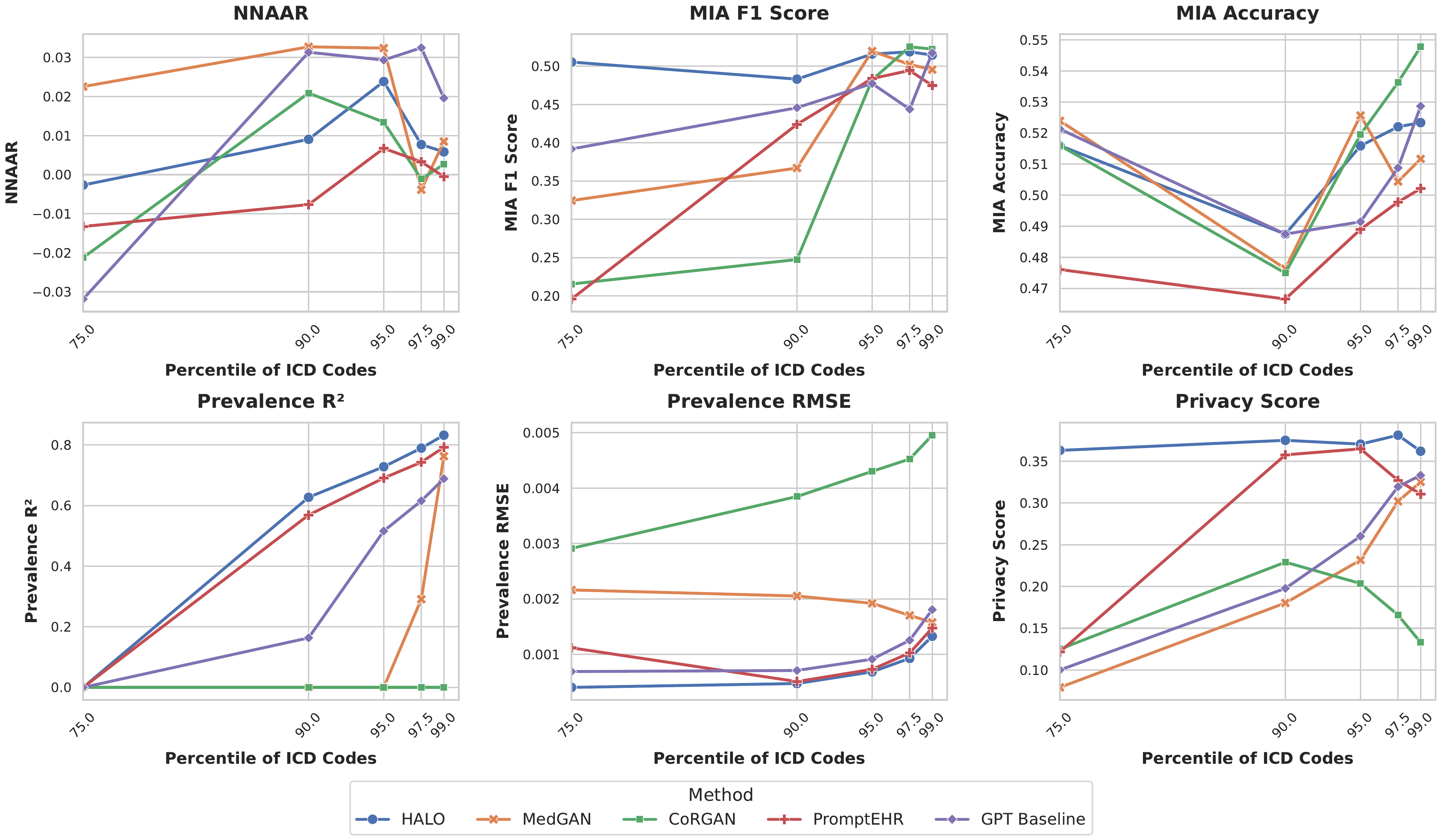}
\caption{Figure 2: Impact of Long-Tailed Code Distributions on Generative Utility and Privacy.}
\label{fig:longtailed}
\end{figure*}

We perform an ablation study evaluating model performance across increasing percentiles of the ICD code vocabulary, ranging from the top 75\% (most frequent) to the 99\% (including rare long-tail codes) by dropping top percentile codes from the real vs synthetic EHRs. From Figure~\ref{fig:longtailed}, we see critical insights into how generative models handle the heterogeneity of Electronic Health Record (EHR) data, specifically the high-cardinality and long-tailed nature of medical codes.

As the scope of modeled codes expands from the 75th to the 99th percentile, the HALO (blue) and PromptEHR (red) models demonstrate superior capacity to capture the underlying data distribution. Both models show a consistent increase in Prevalence $R^2$, reaching nearly 0.8 at the 99th percentile. This indicates that these architectures successfully model the complex dependencies of rare codes rather than treating them as noise. In contrast, CorGAN (green) exhibits a fundamental failure mode. Its Prevalence $R^2$ remains effectively flat at 0.0 across all percentiles, and its RMSE spikes dramatically at the 99th percentile, suggesting it collapses to generating only the most frequent modes or fails to learn the sparse vector space entirely. This is corroborated in Table~\ref{tab:unique_codes}, with CorGAN generating the least number of Unique Codes. 

Interestingly, the GPT baseline also generates fewer unique codes than the other sequential methods, but still performs well for more common codes, as shown by its competitive performance in Table~\ref{tab:synthetic_eval}. The GPT Baseline (purple) performs competitively up to the 97.5th percentile but suffers a sharp degradation in Prevalence $R^2$ at the 99th percentile, highlighting the necessity of specialized algorithms like HALO to stabilize generation in the extreme tail.

The Membership Inference Attack (MIA) metrics confirm that rare codes act as quasi-identifiers. As the analysis moves into the long tail (90th $\rightarrow$ 99th percentile), MIA Accuracy and F1 Scores rise for nearly all models. The long-tail ablation demonstrates that while token-based autoregressive models (PromptEHR, GPT) are best suited for capturing the full breadth of medical history, the inclusion of rare codes introduces significant privacy risks that must be mitigated. Flattened methods like CorGAN and MedGAN are shown to be insufficient for high-fidelity synthesis of complex, long-tailed EHR data.

We complement this percentile-based ablation with several orthogonal views of the same phenomenon in Appendix~\ref{sec:appendix_marginals}. The rank-frequency plot (Figure~\ref{fig:appendix_zipf}) shows that real MIMIC-III spans five decades of prevalence while all synthetic samples truncate well before the rare tail. Coverage of the real vocabulary ranges from 55\% (HALO) down to 21\% (CorGAN) in a 10k-patient sample (Table~\ref{tab:appendix_coverage}), which the aggregate Prevalence $R^2$ reported in Table~\ref{tab:synthetic_eval} almost entirely hides. The per-bucket decomposition (Figure~\ref{fig:appendix_bucket_r2}) makes this explicit: every model's per-decile $R^2$ is strong in the top decile and degrades toward zero in the rarest deciles, with CorGAN collapsing to near-zero by Q7.

\subsection{Discussion}
In this work, we confront the practical reproducibility gap in synthetic EHR generation not by proposing yet another generator, but by building the end-to-end benchmarking substrate that the field has been missing. Our framework ties three pieces together under a single pipeline---\textit{Data $\rightarrow$ Standardized Model $\rightarrow$ Standardized Training $\rightarrow$ Unified Evaluation}---standardizing disparate baselines.
Along the way, we make concrete contributions at each stage of that pipeline. 

At the data and model layers, we restore canonical baselines to full ICD-9 granularity, resurrecting MedGAN's count-matrix mode that was lost in successive PyTorch ports, extending CorGAN's convolutional architecture to natively handle full ICD codes, and providing fully working re-implementations of PromptEHR and HALO. At the same time, we add a deliberately lightweight GPT-2 baseline from the general-purpose sequence-modeling literature---one that turns out to be remarkably competitive and forces specialized architectures to justify their added complexity. At the evaluation layer, we introduce an architecture-agnostic privacy-utility suite that applies identically to GAN- and transformer-based generators, report bootstrapped confidence intervals, and expose the long-tailed failure modes that truncated benchmarks hide.

Taken together, these contributions surface a consistent empirical picture: flat models capture disease marginals almost perfectly but collapse on the joint distribution and the long tail, while sequential models sacrifice a small amount of marginal alignment in exchange for substantially better distributional fidelity, and all evaluated models today remain conservative on privacy. Crucially, the results are now completely reproducible by our training and evaluation scripts. Appendix~\ref{sec:appendix_ablations} supports each of these claims with extended marginal, structural, chapter-level, and co-occurrence ablations computed directly from the released 10k-patient synthetic samples.
We view this framework as a living substrate rather than a finished artifact. Our current instantiation is intentionally scoped to longitudinal ICD codes on MIMIC-III, but each layer of the pipeline is decoupled from that choice, and we actively invite community contributions: new generators, new modalities (labs, vitals, medications, demographics), migration to ICD-10, PheCodes, and DRG codes, and stronger privacy and downstream-utility metrics. Our hope is not that this work is the ``final'' benchmark, but that it lowers the engineering barrier enough that the next generation of synthetic EHR research can focus on the science rather than on re-implementing each other's code.

\subsection{Limitations}

While our framework advances the standardization of EHR synthesis, several critical challenges remain that point toward high-impact directions for future research.
\textbf{Additional Synthetic EHR Generation Methods:} While this work serves as a first step towards truly democratizing and comparing many ML methods, there remain various baselines that were initially too time-consuming to reimplement E.g. EHRDiff \citep{yuan2023ehrdiff} ConSequence \citep{theodorou2024consequence}, and more. Of note, methods that focus on privacy e.g. EHRSafe \citep{yoon2023ehr} and TimeDiff \citep{tian2024reliable} remain top of mind for future additions to the public codebase. By providing a starting point, we hope to welcome community contributions of missing methods/evaluations as well, to truly encourage the open source community to collectively reduce the burden of future EHR benchmarking.
\textbf{High-Dimensional Sparsity:}
A pervasive limitation in current EHR generation research is the inadequate handling of the long-tailed distribution of medical codes. Similarly, this makes benchmarking generic tabular synthesizers difficult \cite{patki2016synthetic}. Future work must move beyond these simplifications to develop architectures capable of modeling the full, untruncated vocabulary, thereby confronting the inherent sparsity of real-world clinical data rather than circumventing it.
\textbf{High-Dimensional Evaluations:}
Unlike other domains where data is regular, in EHR generation, metrics need to be universally applicable across multiple models as well as scalable across a large number of codes and visits. While precision and recall metrics based on manifold estimation (e.g., $\alpha$-precision, $\beta$-recall \citep{alaa2022faithful}) offer theoretical rigor, they rely on exact nearest-neighbor calculations that become computationally intractable \citep{shi2024tabdiff, garuti2025diffusion} in the high-dimensional, potentially sequential feature space of EHRs. Developing scalable, approximation-based metrics that can reliably assess distribution coverage in high dimensions remains an open problem.
\textbf{Beyond MIMIC-III and ICD-9:}
The field's heavy reliance on the MIMIC-III dataset has inadvertently anchored benchmarking efforts to the ICD-9 coding system, which was deprecated in 2015. Modern healthcare systems have long since migrated to ICD-10 \citep{world2004international}, which offers significantly greater granularity and flexibility, including codes for Social Determinants of Health (SDOH). Similarly, there exists other types of commonly used codes such as PheCodes \citep{wu2019mapping}, Diagnosis-Related Group (DRG) codes \citep{centers2019icd, wang2024drg} The continued optimization of models for an obsolete coding standard limits the translational utility of synthetic data. We urge the community to pivot benchmarking efforts toward flexible implementations that can adapt across multiple types of codes, ensuring that synthetic EHRs remain relevant to contemporary medical practice.



\bibliography{jmlr-sample}

\newpage
\appendix

\section{Author Contributions}
Author1 led implementation of CorGAN, MedGAN, and PromptEHR baselines.
Author2 led ideation, evaluation.
Author3 led the implementation of the HALO baseline.
Author4 led example creation, code cleanup, and PR reviews.
Author5 contributed overall project and computational support.

\section{Data and Code Availability}
This paper uses the MIMIC-III dataset
\citep{johnson2016mimic}, which is available on the PhysioNet repository
\citep{PhysioNet-mimiciii-1.4}.
The temporarily anonymized code will be available at \url{https://anonymous.4open.science/r/reproducible_synthetic_ehr-DC11/}. 
\paragraph*{Institutional Review Board (IRB)}
We did not require IRB approval, as all research was performed on publicly available data using open-source software.
must state this to be the case.

\section{Additional Discussion}

\subsection{Extending the Framework Beyond ICD Codes}
\label{sec:extensibility}
A central reviewer concern is whether this framework is useful only for ICD diagnosis generation, or whether it can serve as a substrate for richer synthetic EHR work. By design, each stage of our pipeline is decoupled from the ICD-9 vocabulary and can be extended independently.

\paragraph{Data layer.} Our long-format schema (\texttt{subject\_id}, \texttt{visit\_id}, \texttt{code}) does not assume that \texttt{code} is an ICD diagnosis. Any categorical code system (ICD-10, PheCodes, DRG codes, ATC medication codes, or CPT procedure codes) plugs in without schema changes. Additional modalities (labs, vitals, static demographics) can be added as extra columns or as parallel long-format tables keyed on \texttt{(subject\_id, visit\_id)}. Because this layout is aligned with the MEDS convention~\citep{meds2024}, any upstream loader that emits MEDS-compatible data is usable out of the box.

\paragraph{Model layer.} Each benchmarked model is registered behind a common PyHealth-style interface that exposes \texttt{fit(dataset)} and \texttt{generate(n)}. Adding a new generator---or re-running an existing one on a new code system---requires only that the model emit records in the long-format schema above. None of the training loops, evaluation scripts, or bootstrapping code are specific to ICD-9 beyond the vocabulary size.

\paragraph{Evaluation layer.} Because the metric suite operates on the final discretized records rather than on model-internal quantities, extending evaluation to new modalities reduces to extending the underlying record type. For continuous modalities (labs, vitals), the prevalence metric generalizes to marginal distribution comparisons (e.g., Wasserstein-1 per feature), the discriminator-based fidelity score applies unchanged, and the set-distance privacy metric generalizes naturally to records that concatenate categorical and continuous features. Our goal is for this framework to serve as the default harness for such contributions rather than to enumerate every possible modality in this initial release.

\subsection{Computational Cost of Full-Vocabulary Training}
\label{sec:compute}
Scaling to the full ICD-9 vocabulary meaningfully increases training cost relative to 3-digit-truncated baselines, and the cost is highly uneven across architectures. PromptEHR is the most expensive model in our suite at approximately 16 hours on a single A100/H100 GPU (64GB) for 20 epochs over 45{,}520 patients, driven by its BART-style encoder-decoder and 512-token maximum sequence length. HALO trains in under 6 hours on the same hardware with its 12-layer GPT-2-style backbone. The simple GPT-2 baseline is by far the cheapest: with 8 layers, embedding dimension 512, and our word-level tokenizer, it converges in roughly 2 hours. The flattened GAN baselines (MedGAN, CorGAN) train in under 1 hour each, because their linear and convolutional architectures scale with vocabulary size and are bottlenecked by adversarial training dynamics rather than sequence length. In practice, PromptEHR's 16-hour figure is likely overkill for our setting; we report it because it is the configuration we actually used, and we highlight it to clarify that the ``high cost'' of full-vocabulary training is dominated by a single heavyweight baseline, not by the framework itself. GPU memory stayed below 40GB for every model at full ICD-9 resolution.

\subsection{Downstream Utility of Full-Granularity Codes}
\label{sec:downstream}
A reasonable question is whether modeling all 6{,}955 ICD-9 codes is worth the extra complexity over the 1{,}071 three-digit buckets used by prior baselines. We argue that truncation silently destroys clinical content that downstream users cannot recover. Under the three-digit code 250 (``Diabetes mellitus''), a well-controlled Type 2 diabetic is indistinguishable from a patient in Type 1 diabetic ketoacidosis (250.11); under 427 (``Cardiac dysrhythmias''), benign sinus tachycardia and life-threatening ventricular fibrillation collapse into the same token. Any model trained on truncated codes inherits this collapse, which makes the resulting synthetic data unusable for precision-medicine risk scoring, phenotype-specific cohort construction, or outcome prediction for severe subtypes---the exact downstream tasks that synthetic EHR data is meant to enable. Fully quantifying the gap in downstream utility (e.g., via readmission prediction conditioned on full-granularity synthetic data) requires models that jointly emit diagnosis codes and the static demographics needed for downstream labels, which remains future work we are actively adding to the public codebase.

\section{Additional Tables}

\begin{table}[ht]
\centering
\caption{Comparison of Unique Codes in Synthetic EHR Datasets (10k Samples)     \label{tab:unique_codes}}
\begin{tabular}{lc}
\toprule
\textbf{Generation Method} & \textbf{Unique Codes} \\
\midrule
HALO & 3,849 \\
MedGAN & 2,429 \\
GPT Baseline & 2,244 \\
PromptEHR & 1,772 \\
CorGAN & 1,467 \\
\bottomrule
\end{tabular}
\end{table}

\section{Machine Learning Models for Discriminative Score}
\label{sec:appendix_discriminator_eval}
\subsection{Sequential Discriminator (LSTM)}
For the sequential generative models (HALO and PromptEHR), capturing the temporal dynamics and dependencies between subsequent visits is paramount. A flat classifier would fail to detect unnatural temporal transitions (e.g., a diagnosis appearing out of clinical order). Therefore, we employed a Long Short-Term Memory (LSTM) network as the discriminator. 

The LSTM discriminator ingests the patient visit sequence $x = (v_1, v_2, \dots, v_T)$ where each visit $v_t$ is a multi-hot vector of diagnosis codes. The model is trained to minimize the binary cross-entropy loss, classifying samples as either \textit{Real} ($y=1$) or \textit{Synthetic} ($y=0$). High prediction accuracy by the LSTM indicates that the synthetic generator failed to capture complex temporal dependencies, rendering the data easily distinguishable.

\subsection{Tabular Discriminator (Random Forest)}
For the flat generative baselines (MedGAN and CorGAN), which generate aggregated patient records or single-snapshot visits, temporal dependencies are not the primary feature of interest. Instead, the correlation structure between features (co-occurrence of codes) is the critical evaluative criterion.

We employed a Random Forest (RF) classifier as the discriminator for these models. Random Forests are robust against overfitting and highly effective at capturing non-linear interactions in high-dimensional tabular data. The input to the RF was the aggregated multi-hot vector representation of patient history. Similar to the LSTM approach, the RF was trained to distinguish real from synthetic records. A high classification accuracy implies that the synthetic data resides in a distinct region of the feature space compared to the real distribution.

\section{Hyperparameters Used}
\label{appendix:hyperparams}

\subsection{MedGAN Architecture and Training Configuration}

\paragraph*{MedGAN Architecture.}

MedGAN employs a linear layer architecture with separate encoder and decoder networks. The encoder compresses binary patient records $x \in \{0,1\}^d$ to latent codes $z \in \mathbb{R}^h$ where $h$ is the hidden dimension. The
decoder reverses this mapping with activation functions determined by data mode: sigmoid for binary mode (outputs in $[0,1]$ interpreted as probabilities), ReLU or Softplus for count mode (outputs in $[0,\infty)$ interpreted as
visit counts). The generator network $G_\phi: \mathbb{R}^{128} \rightarrow \mathbb{R}^h$ transforms noise vectors to synthetic latent codes, employing residual connections for improved gradient flow. The discriminator $D_\psi:
\mathbb{R}^h \rightarrow [0,1]$ uses minibatch averaging to prevent mode collapse.

\paragraph*{MedGAN Training.}

\begin{table}[ht]
\centering
\caption{MedGAN Training Hyperparameters}
\label{tab:MedGAN-train}
\begin{tabular}{ll}
\toprule
\textbf{Parameter} & \textbf{Value} \\
\midrule
Vocabulary size ($d$) & 6955 (full ICD-9) \\
Hidden dimension ($h$) & 128 \\
Latent dimension & 128 \\
\midrule
\multicolumn{2}{l}{\textit{Autoencoder Pretraining}} \\
\quad Epochs & 100 \\
\quad Loss (binary mode) & Binary cross-entropy \\
\quad Loss (count mode) & Mean squared error \\
\quad Optimizer & Adam \\
\quad Learning rate & 0.001 \\
\quad Momentum $\beta_1$ & 0.9 \\
\quad Momentum $\beta_2$ & 0.999 \\
\quad Weight decay & 0.0001 \\
\midrule
\multicolumn{2}{l}{\textit{Adversarial Training}} \\
\quad Epochs & 1000 \\
\quad Batch size & 128 \\
\quad Optimizer & Adam \\
\quad Learning rate & 0.001 \\
\quad Weight decay & 0.0001 \\
\quad Generator & Residual connections \\
\quad Discriminator & Minibatch averaging \\
\bottomrule
\end{tabular}
\end{table}

\paragraph*{MedGAN Generation.}

Generation differs by mode. \textit{Binary mode:} Sample $\mathbf{z} \sim \mathcal{N}(0, \mathbf{I}_{128})$, generate latent $\mathbf{h} = G(\mathbf{z})$, decode with sigmoid $\mathbf{\hat{x}} = \sigma(D(\mathbf{h}))$, threshold
at 0.5: $x_i = \mathbb{1}[\hat{x}_i \geq 0.5]$. \textit{Count mode:} Sample $\mathbf{z}$, generate $\mathbf{h} = G(\mathbf{z})$, decode with ReLU $\mathbf{\hat{x}} = \text{ReLU}(D(\mathbf{h}))$, round to integers and clip: $x_i
= \text{clip}(\text{round}(\hat{x}_i), 0, \max_{\text{train}})$.

\paragraph*{MedGAN Count Mode Restoration.}

The original MedGAN (Choi et al., 2017) supported both binary and count modes. The reference PyTorch implementation from the CorGAN codebase restricted MedGAN to binary mode only, eliminating the capability to model visit
frequencies (e.g., distinguishing one versus five visits for a condition). SynthEHRella adopted this version. We restore count mode via conditional decoder activation based on \texttt{data\_mode} parameter. Binary mode uses
sigmoid activation $\rightarrow [0,1]$; count mode uses ReLU $\rightarrow [0,\infty)$. Implementation: final decoder layer applies \texttt{nn.Sigmoid()} for binary mode, \texttt{nn.ReLU()} for count mode.

\subsection{CorGAN Architecture and Training Configuration}

\paragraph*{CorGAN Architecture.}

CorGAN employs 1D convolutional networks for encoder and decoder. The encoder applies eight Conv1d layers with LeakyReLU($\alpha=0.2$) activations (final layer uses Tanh) to compress binary records $x \in \{0,1\}^{6955}$ to
scalar latent $z \in [-1,1]$. The decoder reverses this via eight ConvTranspose1d layers with ReLU activations (final layer uses Sigmoid) to produce $\hat{x} \in [0,1]^{6955}$. The generator $G_\phi: \mathbb{R}^{128} \rightarrow
\mathbb{R}$ uses three fully-connected layers with LeakyReLU and final Tanh. The discriminator $D_\psi: \mathbb{R} \rightarrow [0,1]$ uses four convolutional layers. All layers use Xavier uniform initialization.

\begin{table}[ht]
\centering
\caption{CorGAN 8-Layer Architecture Specification}
\label{tab:CorGAN-arch-detail}
\begin{tabular}{llll}
\toprule
\textbf{Layer} & \textbf{Type} & \textbf{Dimension} & \textbf{Activation} \\
\midrule
\multicolumn{4}{l}{\textit{Encoder (Conv1d)}} \\
1 & Conv1d & 6955 $\rightarrow$ 3476 & LeakyReLU(0.2) \\
2 & Conv1d & 3476 $\rightarrow$ 1736 & LeakyReLU(0.2) \\
3 & Conv1d & 1736 $\rightarrow$ 578 & LeakyReLU(0.2) \\
4 & Conv1d & 578 $\rightarrow$ 192 & LeakyReLU(0.2) \\
5 & Conv1d & 192 $\rightarrow$ 63 & LeakyReLU(0.2) \\
6 & Conv1d & 63 $\rightarrow$ 20 & LeakyReLU(0.2) \\
7 & Conv1d & 20 $\rightarrow$ 4 & LeakyReLU(0.2) \\
8 & Conv1d & 4 $\rightarrow$ 1 & Tanh \\
\midrule
\multicolumn{4}{l}{\textit{Decoder (ConvTranspose1d)}} \\
1 & ConvTranspose1d & 1 $\rightarrow$ 4 & ReLU \\
2 & ConvTranspose1d & 4 $\rightarrow$ 20 & ReLU \\
3 & ConvTranspose1d & 20 $\rightarrow$ 63 & ReLU \\
4 & ConvTranspose1d & 63 $\rightarrow$ 192 & ReLU \\
5 & ConvTranspose1d & 192 $\rightarrow$ 578 & ReLU \\
6 & ConvTranspose1d & 578 $\rightarrow$ 1736 & ReLU \\
7 & ConvTranspose1d & 1736 $\rightarrow$ 3476 & ReLU \\
8 & ConvTranspose1d & 3476 $\rightarrow$ 6955 & Sigmoid \\
\midrule
\multicolumn{4}{l}{\textit{Generator (FC)}} \\
Input & Linear & 128 $\rightarrow$ 256 & LeakyReLU \\
Hidden & Linear & 256 $\rightarrow$ 128 & LeakyReLU \\
Output & Linear & 128 $\rightarrow$ 1 & Tanh \\
\bottomrule
\end{tabular}
\end{table}

\paragraph*{CorGAN Training.}

\begin{table}[ht]
\centering
\caption{CorGAN Training Hyperparameters}
\label{tab:CorGAN-train-detail}
\begin{tabular}{ll}
\toprule
\textbf{Parameter} & \textbf{Value} \\
\midrule
Vocabulary size ($d$) & 6955 (full ICD-9) \\
Latent dimension & 128 \\
\midrule
\multicolumn{2}{l}{\textit{Autoencoder Pretraining}} \\
\quad Epochs & 10 (optimal configuration) \\
\quad Loss & Binary cross-entropy \\
\quad Formula & $-\frac{1}{d}\sum_{j=1}^d [x_j \log \hat{x}_j + (1-x_j)\log(1-\hat{x}_j)]$ \\
\quad Optimizer & Adam \\
\quad Learning rate & 0.001 \\
\quad Momentum $\beta_1$ & 0.5 \\
\quad Momentum $\beta_2$ & 0.9 \\
\midrule
\multicolumn{2}{l}{\textit{Adversarial Training (WGAN-GP)}} \\
\quad Epochs & 500 \\
\quad Batch size & 256 (optimal configuration) \\
\quad Optimizer & Adam \\
\quad Learning rate & 0.001 \\
\quad Discriminator:Generator ratio & 5:1 \\
\quad Gradient penalty $\lambda_{\text{GP}}$ & 10 \\
\quad Weight clipping & $[-0.01, 0.01]$ \\
\quad Initialization & Xavier uniform \\
\bottomrule
\end{tabular}
\end{table}

\textit{WGAN-GP Objective.} Discriminator minimizes $\mathcal{L}_D = \mathbb{E}_{z}[D(G(z))] - \mathbb{E}_{x}[D(\text{Enc}(x))] + \lambda_{\text{GP}} \mathbb{E}_{\hat{z}}[(\|\nabla_{\hat{z}} D(\hat{z})\|_2 - 1)^2]$ where
$\hat{z} = \epsilon \text{Enc}(x) + (1-\epsilon)G(z)$ with $\epsilon \sim \text{Uniform}(0,1)$. Generator minimizes $\mathcal{L}_G = -\mathbb{E}_{z}[D(G(z))]$.

\paragraph*{CorGAN 8-Layer Architecture Development.}

The original 6-layer architecture (1071 $\rightarrow$ 534 $\rightarrow$ 265 $\rightarrow$ 87 $\rightarrow$ 28 $\rightarrow$ 8 $\rightarrow$ 1) used fixed kernel sizes and strides. Extending to 6,955 codes required solving
dimension-matching: CNNs cannot arbitrarily resize outputs without architectural changes. Extended to 8 layers with dimension pathway 6955 $\rightarrow$ 3476 $\rightarrow$ 1736 $\rightarrow$ 578
$\rightarrow$ 192 $\rightarrow$ 63 $\rightarrow$ 20 $\rightarrow$ 4 $\rightarrow$ 1, enabling native dimension matching without pooling artifacts.

\paragraph*{Implementation History and Modifications.}

\textit{CorGAN:} Original \citep{torfi2020CorGAN} used 6-layer CNN for 1,071 three-digit ICD-9 codes. SynthEHRella ported to PyTorch, maintaining 6 layers and 1,071 codes. Our implementation: (1) corrected stride bug in decoder
layer 3 (stride 2 $\rightarrow$ 4), (2) extended to 8 layers supporting 6,955 codes. \textit{MedGAN:} Original \citep{choi2017generating} supported binary and count modes for 1,071 codes via linear layers. The reference PyTorch
implementation from the CorGAN codebase removed count mode, restricting to binary only. SynthEHRella adopted this version. Our implementation: (1) restored count mode via conditional decoder activation, (2) extended to 6,955
codes (trivial for linear architecture).

\subsection{PromptEHR Architecture and Training Configuration}
\paragraph*{Model Architecture.}

PromptEHR employs a modified BART architecture for conditional EHR generation. The model consists of a bidirectional transformer encoder $\text{Enc}_\phi$ processing demographic prompt embeddings $p$ to produce contextualized
representations $H_{\text{enc}} = \text{Enc}_\phi(p)$, and an auto-regressive transformer decoder $\text{Dec}_\psi$ generating medical event sequences. The conditional generation probability factorizes as:

\begin{equation}
P(x_{1:T} \mid p) = \prod_{t=1}^{T} P(x_t \mid x_{<t}, H_{\text{enc}})
\end{equation}

where $x_{1:T} = \{v_1, v_2, \ldots, v_T\}$ represents a sequence of $T$ clinical visits.

\textit{Hierarchical Tokenization.} We represent the two-level EHR hierarchy (patient timeline $\rightarrow$ visits $\rightarrow$ codes) using special delimiter tokens $\langle \text{v} \rangle$ and $\langle \text{/v} \rangle$:
$x = [\langle \text{v} \rangle, c_{1,1}, \ldots, c_{1,n_1}, \langle \text{/v} \rangle, \langle \text{v} \rangle, c_{2,1}, \ldots]$. The vocabulary $\mathcal{V}_{\text{full}} = \mathcal{V}_{\text{med}} \cup
\mathcal{V}_{\text{special}}$ contains ICD-9 diagnosis codes and seven special tokens ($\langle \text{pad} \rangle, \langle \text{s} \rangle, \langle \text{/s} \rangle, \langle \text{unk} \rangle, \langle \text{v} \rangle,
\langle \text{/v} \rangle, \langle \text{mask} \rangle$) assigned IDs 0--6, with diagnosis codes beginning at ID 7.

\paragraph*{Demographic Conditioning.}

Patient demographics (age, sex) are encoded via learned prompt embeddings with reparameterization, projecting raw values into BART hidden space ($d=768$). For age (continuous), a numerical encoder applies:

\begin{equation}
h_{\text{num}} = W_{\text{proj}}(W_{\text{age}} \cdot \text{age} + b_{\text{age}})
\end{equation}

where $W_{\text{age}} \in \mathbb{R}^{1 \times 128}$, $b_{\text{age}} \in \mathbb{R}^{1 \times 128}$, and $W_{\text{proj}} \in \mathbb{R}^{128 \times 768}$. For sex (categorical), a learned embedding table ($2 \times 128$)
encodes male/female, adds a feature-specific bias, and projects to dimension 768. Final conditioning: $p = [h_{\text{age}}, h_{\text{sex}}] \in \mathbb{R}^{2 \times 768}$, prepended to encoder input.

\paragraph*{Training Details.}

\begin{table}[ht]
\centering
\caption{PromptEHR Training Hyperparameters}
\label{tab:PromptEHR-train}
\begin{tabular}{ll}
\toprule
\textbf{Parameter} & \textbf{Value} \\
\midrule
Training set size & 45,520 patients ($\mathcal{D}_{\text{train}}$) \\
Holdout set size & 1,000 patients ($\mathcal{D}_{\text{holdout}}$) \\
Objective & $-\frac{1}{N} \sum_{i=1}^{N} \log P(x_i \mid p_i; \phi, \psi)$ \\
Optimizer & AdamW \\
Learning rate ($\eta$) & $1 \times 10^{-5}$ \\
Weight decay ($\lambda$) & 0.01 \\
Batch size & 16 sequences per GPU \\
Max sequence length & 512 tokens \\
Warmup steps & 1,000 \\
Training schedule & Linear warmup + linear decay \\
Total epochs & 20 \\
Early stopping metric & Validation perplexity \\
Training time & $\approx$16 hours (A100/H100 GPU, 64GB) \\
\midrule
\multicolumn{2}{l}{\textit{Data Augmentation}} \\
\quad Mask infilling & Enabled \\
\quad Token deletion & Enabled \\
\quad Token replacement ($p_{\text{replace}}$) & 0 \\
\bottomrule
\end{tabular}
\end{table}

\textit{Token Replacement and Frequency Preservation.} We disable token replacement ($p_{\text{replace}} = 0$) to preserve the empirical code frequency distribution. Stochastic token replacement, where diagnosis codes are
randomly replaced with other codes during augmentation, inverts frequency distributions in long-tailed medical vocabularies: rare codes (occurring $<$10 times) are overrepresented by up to 4,700$\times$ while common codes are
depleted, as uniform random replacement disproportionately introduces rare codes. We retain mask infilling and token deletion, which preserve relative frequencies.

\paragraph*{Generation Parameters.}

\begin{table}[ht]
\centering
\caption{PromptEHR Generation Hyperparameters}
\label{tab:PromptEHR-gen}
\begin{tabular}{ll}
\toprule
\textbf{Parameter} & \textbf{Value} \\
\midrule
Decoding strategy & Nucleus sampling (top-$p$) \\
Top-$p$ & 0.95 \\
Top-$k$ & 0 (disabled) \\
Temperature ($\tau$) & 1.0 \\
Frequency guidance ($\alpha$) & 2.0 \\
Decoder start token & $\langle \text{/s} \rangle$ (ID 2) \\
\midrule
\multicolumn{2}{l}{\textit{Structure Sampling}} \\
\quad Demographics & Sampled from $\mathcal{D}_{\text{train}}$ empirical distribution \\
\quad Number of visits & Sampled from pooled visit count distribution \\
\quad Codes per visit & Sampled from pooled visit length distribution \\
\bottomrule
\end{tabular}
\end{table}

We employ nucleus sampling with $p = 0.95$, renormalizing over the nucleus $\mathcal{V}_p$ (smallest subset where $\sum_{x' \in \mathcal{V}_p} P(x' \mid x_{<t}) \geq p$). Top-$k$ is disabled ($k=0$) to preserve access to the
full vocabulary ($|\mathcal{V}_{\text{full}}| \approx 5000$ tokens), preventing exclusion of rare codes in the long-tail distribution.

\paragraph*{Frequency-Guided Generation.}

The function \texttt{build\_frequency\_prior()} constructs a log-frequency tensor $f \in \mathbb{R}^{|\mathcal{V}_{\text{full}}|}$ where $f_c = \log(P_{\text{train}}(c) + \epsilon)$ for each code $c$, with $\epsilon = 10^{-10}$.
Special tokens receive neutral prior ($f_s = 0$). During generation, model logits are blended via $\ell_{\text{guided}} = \ell_{\text{model}} + \alpha \cdot f$, applied post-training without modifying model weights.

The function \texttt{build\_first\_code\_prior()} constructs demographic-conditioned distributions $P(c_{\text{first}} \mid \text{age\_bin}, \text{sex})$ by stratifying first diagnosis codes by age (9 bins: 0--10, 10--20,
\ldots, 80--90) and sex.

\paragraph*{Post-Processing and Framework Integration.}

Generated sequences are post-processed by splitting at $\langle \text{/v} \rangle$ delimiters, extracting codes, assigning time indices, and converting to wide format (one row per patient) and long format (one row per visit).

\subsection{HALO Configuration}
\paragraph*{Model Architecture}
HALO uses a 12-layer transformer encoder with 768-dimensional embeddings and 12 attention heads, following the GPT-2 architecture. The model processes sequential EHR visits through a CoarseTransformerModel backbone with
multi-head self-attention and position embeddings. An autoregressive head (FineAutoregressiveHead) generates medical codes sequentially within each visit using masked linear layers that enforce causal dependencies.
Table~\ref{tab:halo_training_params} details the complete architecture and training configuration.

\begin{table}[ht]
\centering
\caption{HALO Training Hyperparameters}
\label{tab:halo_training_params}
\resizebox{\linewidth}{!}{
\begin{tabular}{lll}
\toprule
\textbf{Parameter} & \textbf{Value} & \textbf{Description} \\
\midrule
\multicolumn{3}{l}{\textit{Model Architecture}} \\
\texttt{n\_embd} & 768 & Embedding dimension \\
\texttt{n\_layer} & 12 & Number of transformer layers \\
\texttt{n\_head} & 12 & Number of attention heads \\
\texttt{n\_positions} & 56 & Maximum sequence length \\
\texttt{n\_ctx} & 48 & Context window size \\
\texttt{layer\_norm\_epsilon} & $10^{-5}$ & Layer normalization epsilon \\
\texttt{initializer\_range} & 0.02 & Weight initialization range \\
\midrule
\multicolumn{3}{l}{\textit{Vocabulary}} \\
\texttt{code\_vocab\_size} & 6,841 & Medical diagnosis codes \\
\texttt{label\_vocab\_size} & 25 & HCUP CCS clinical groupings \\
\texttt{special\_vocab\_size} & 3 & Special tokens (start, end, padding) \\
\texttt{total\_vocab\_size} & 6,869 & Sum of all vocabulary components \\
\midrule
\multicolumn{3}{l}{\textit{Training}} \\
\texttt{batch\_size} & 48 & Training batch size \\
\texttt{epoch} & 80 & Number of training epochs \\
\texttt{lr} & $10^{-4}$ & Learning rate (Adam optimizer) \\
\texttt{pos\_loss\_weight} & None & Positive class weighting (disabled) \\
Validation frequency & Every 500 batches & Checkpoint evaluation interval \\
Loss function & Weighted BCE & Binary cross-entropy with masking \\
\midrule
\multicolumn{3}{l}{\textit{Data Split}} \\
Train & 72\% & 80\% of data, then 90\% for train \\
Validation & 8\% & 80\% of data, then 10\% for validation \\
Test & 20\% & Held-out test set \\
\bottomrule
\end{tabular}
}
\end{table}

\paragraph*{Vocabulary Construction}
The reference implementation constructs vocabulary exclusively from diagnosis codes flagged as \texttt{use\_in\_benchmark=True} in the HCUP CCS taxonomy, filtering codes before training. Our implementation constructs vocabulary
from all diagnosis codes observed in the training data. The HCUP benchmark taxonomy is used to create patient-level labels for clinical groupings but does not restrict the code vocabulary. Total vocabulary size is computed
dynamically as $|V_{\text{total}}| = |V_{\text{codes}}| + |V_{\text{labels}}| + 3$ where special tokens include start, end, and padding markers. This modification preserves the model architecture while enabling training on the
complete diagnostic code space.

\paragraph*{Training Configuration}
Following the reference implementation, we use batch size 48, learning rate $10^{-4}$, and Adam optimization. Training proceeds for 80 epochs with validation every 500 batches. The model minimizes weighted binary cross-entropy
loss over visit predictions, with optional positive class weighting disabled (pos\_loss\_weight=None).

\paragraph*{Data Representation}
Patient records are encoded as 3D tensors with shape (batch, sequence, vocabulary). Each visit is represented as a multi-hot vector over the vocabulary. Sequence positions 0 and 1 contain start token and patient labels
respectively, positions $2$ to $n+1$ contain visits, position $n+2$ contains the end token, and remaining positions are padded. Masks indicate valid positions for loss computation, excluding padding tokens from gradient updates.

\paragraph*{Generation Procedure}
Synthetic patients are generated autoregressively starting from the start token. At each position, the model predicts a vocabulary-sized logit vector, applies sigmoid activation, and samples codes via Bernoulli sampling.
Generation continues until the model predicts the end token or reaches maximum sequence length (48 positions). Sample batch size is 256 during generation. Table~\ref{tab:halo_generation_params} specifies generation
hyperparameters.

\begin{table}[ht]
\centering
\caption{HALO Generation Hyperparameters}
\label{tab:halo_generation_params}
\resizebox{\linewidth}{!}{
\begin{tabular}{lll}
\toprule
\textbf{Parameter} & \textbf{Value} & \textbf{Description} \\
\midrule
\texttt{sample\_batch\_size} & 256 & Generation batch size \\
\texttt{sample} & True & Stochastic sampling (Bernoulli) \\
Max sequence length & 48 & Maximum visits per patient \\
Sampling strategy & Autoregressive & Sequential visit generation \\
Code activation & Sigmoid + Bernoulli & Per-code probability sampling \\
Early stopping & End token & Generation halts at predicted end token \\
\bottomrule
\end{tabular}
}
\end{table}

\begin{table}[ht]
\centering
\caption{HALO Implementation Notes}
\label{tab:halo-notes}
\begin{tabular}{ll}
\toprule
\textbf{Aspect} & \textbf{Details} \\
\midrule
Implementation source & Theodorou et al.~\citep{theodorou2023synthesize} \\
Modifications & Hard-coded paths updated \\
& Epoch count increased \\
Integration effort & Minimal (out-of-the-box compatibility) \\
\bottomrule
\end{tabular}
\end{table}

\subsection{GPT Baseline}
\begin{table}[ht]
\centering
\caption{Hyperparameters for GPT-2 EHR Baseline \label{tab:gpt_training_params}}
\begin{tabular}{@{}ll@{}}
\toprule
\textbf{Hyperparameter} & \textbf{Value / Range} \\ \midrule
Number of Layers ($n_{layer}$) & 2 -- 8 \\
Embedding Dimension ($n_{embd}$) & 16 -- 512 \\
Number of Heads ($n_{head}$) & 8 \\
Context Window ($n_{ctx}$) & 512 \\
Max Sequence Length & 128 \\
Learning Rate & $1 \times 10^{-4}$  \\
LR Scheduler & Cosine \\
Batch Size & 64 \\
Training Epochs & 50  \\
Optimizer & Adam  \\ \bottomrule
\end{tabular}
\end{table}
The GPT-2 baseline was implemented to evaluate the performance of a standard autoregressive language model on flattened EHR sequences. Unlike the hierarchical structure of HALO, the GPT-2 baseline converts multi-level longitudinal records into a fully one-hot sequential representation. 
The model was trained using the Causal Language Modeling (CLM) objective in Huggingface Trainer \cite{wolf2019huggingface} for 50 epochs with a batch size of 64. We employed the AdamW optimizer with a cosine learning rate scheduler ($\eta_{max}=1\times10^{-4}$). For generation, we utilized nucleus sampling ($top\_p=0.95$) combined with top-k sampling ($k=50$) to ensure diversity while maintaining clinical coherence.

The training process involves the following key steps:
\begin{itemize}
\item \textbf{Data Flattening:} Patient records are transformed from hierarchical visits into linear strings. Each medical ICD code is treated as an individual token, and visits are separated by a special delimiter token, \texttt{VISIT\_DELIM}. 
\item \textbf{Tokenization:} A custom \texttt{WordLevel} tokenizer is trained on the patient sequences to ensure that clinical codes (e.g., "401.9") are treated as atomic units rather than being broken into sub-tokens. Special tokens, including \texttt{[BOS]} (Beginning of Sequence), \texttt{[EOS]} (End of Sequence), and \texttt{[PAD]} (Padding), are added to manage sequence boundaries.
\item \textbf{Causal Language Modeling:} The model is trained using a standard cross-entropy loss to predict the next token in the sequence given all preceding tokens. Due to the high dimensionality of EHR data, the sequence length increases significantly when flattened, necessitating architectural adjustments for memory efficiency.
\end{itemize}

The following table~\ref{tab:gpt_training_params} summarizes the hyperparameter search space and the final configuration used for the GPT-2 baseline experiments. Note that while the standard GPT-2 architecture often utilizes 12 layers, the baseline was frequently constrained to fewer layers (e.g., 3 to 8) to accommodate the memory requirements of expanded EHR sequences.

\section{Additional Ablations and Plots}
\label{sec:appendix_ablations}

To strengthen the empirical picture presented in the main text, we report a broader set of distributional, structural, and code-level ablations. 

\subsection{Dataset-level descriptive statistics}
\label{sec:appendix_descriptive}
Table~\ref{tab:dataset_stats_extra} reports basic descriptive statistics for each synthetic sample alongside the real training split. Flattened baselines (MedGAN, CorGAN) trivially collapse to a single pseudo-visit per patient, so their mean visits/patient is 1.0 by construction; this is the structural limitation that the sequential models (HALO, PromptEHR, GPT) are designed to overcome. We note that PromptEHR comes closest to the real mean codes-per-patient (11.73 vs.\ 12.61), while MedGAN and CorGAN over-fit the per-visit code count because all of a patient's codes are packed into one visit. The sequential models produce shorter visits on average than real MIMIC-III, pointing to a consistent tendency to under-emit codes per encounter that is invisible from aggregate prevalence.

\begin{table}[ht]
\centering
\caption{Descriptive statistics over 10k-patient synthetic samples and the real training split. ``Unique codes'' is the number of distinct ICD-9 codes that ever appear in the dataset. Flattened baselines (MedGAN, CorGAN) produce one pseudo-visit per patient.}
\label{tab:dataset_stats_extra}
\resizebox{\linewidth}{!}{%
\begin{tabular}{lrrrrrr}
\toprule
\textbf{Dataset} & \textbf{Patients} & \textbf{Visits} & \textbf{Visits/patient} & \textbf{Codes/visit} & \textbf{Codes/patient} & \textbf{Unique codes} \\
\midrule
Real (train)  & 45{,}517 & 57{,}683 & 1.27 & 11.04 & 12.61 & 6{,}955 \\
\midrule
HALO          &  9{,}989 & 13{,}216 & 1.32 &  8.17 &  9.71 & 3{,}849 \\
PromptEHR     & 10{,}000 & 12{,}913 & 1.29 &  9.25 & 11.73 & 1{,}772 \\
GPT Baseline  & 10{,}000 & 11{,}583 & 1.16 &  9.15 &  9.90 & 2{,}244 \\
MedGAN        &  9{,}622 &  9{,}622 & 1.00 & 12.52 & 12.52 & 2{,}429 \\
CorGAN        & 10{,}000 & 10{,}000 & 1.00 & 12.56 & 12.56 & 1{,}467 \\
\bottomrule
\end{tabular}%
}
\end{table}

\subsection{Marginal code-frequency ablations}
\label{sec:appendix_marginals}

\begin{figure}[ht]
\centering
\includegraphics[width=\linewidth]{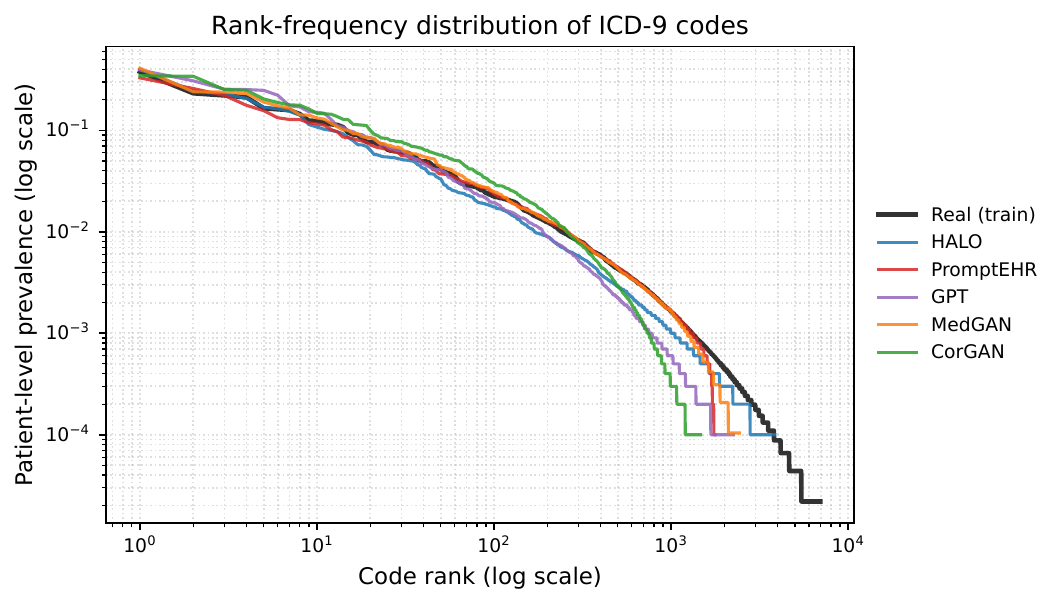}
\caption{Rank--frequency distribution of patient-level ICD-9 prevalence (log--log). HALO best tracks the real long tail; CorGAN and PromptEHR truncate earliest.\label{fig:appendix_zipf}}
\end{figure}

\paragraph{Rank--frequency (Zipf) plot.} Figure~\ref{fig:appendix_zipf} shows the log--log rank--frequency curve of patient-level ICD-9 prevalence for real and synthetic data. Real MIMIC-III exhibits a textbook Zipf-like decay spanning roughly five decades of prevalence, and no synthetic generator recovers the full tail: HALO tracks the real curve furthest into the long tail, while CorGAN and PromptEHR diverge from the real curve by rank $\sim$$10^3$. This is the strongest visual signal that ``rare-code collapse'' is a shared, model-specific failure mode, not a quirk of a single benchmark metric.

\begin{figure}[ht]
\centering
\includegraphics[width=\linewidth]{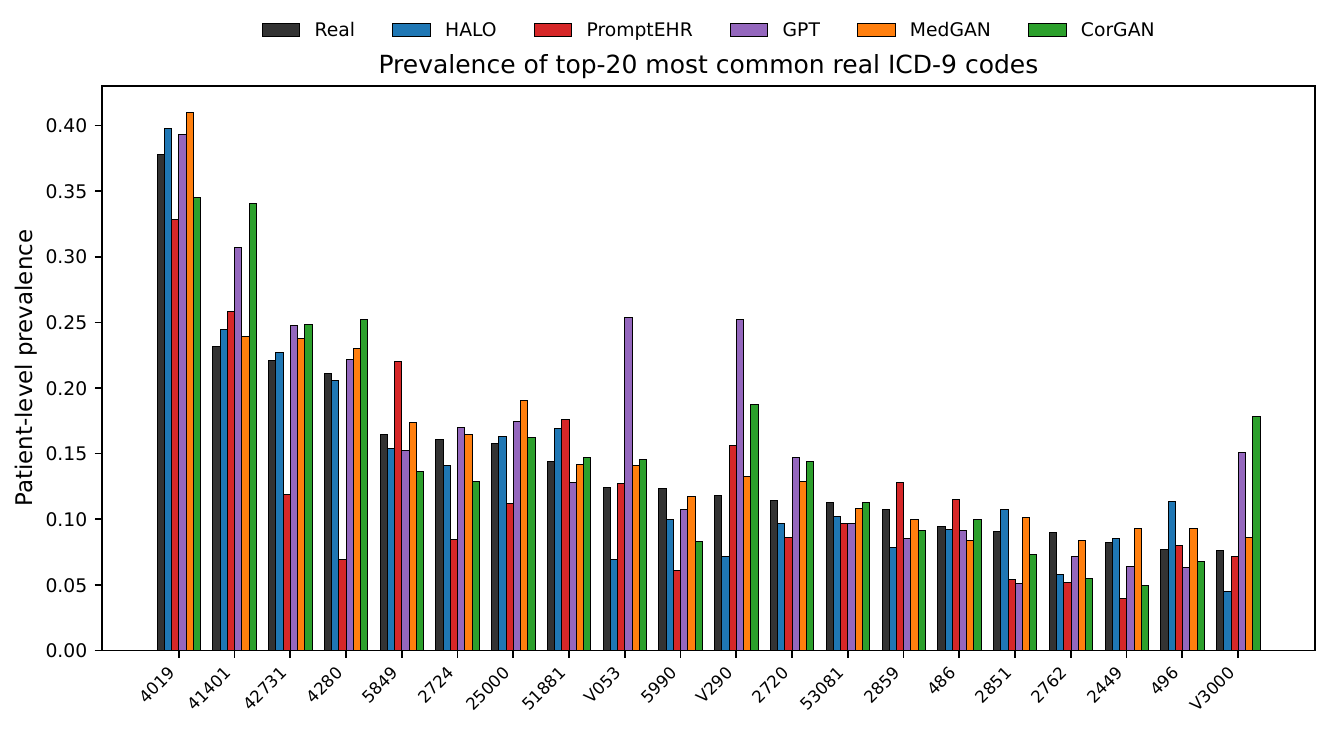}
\caption{Patient-level prevalence of the 20 most common real ICD-9 codes, across each synthetic generator. CorGAN under-predicts even the head of the distribution.\label{fig:appendix_top20}}
\end{figure}

\paragraph{Top-20 code prevalence.} Figure~\ref{fig:appendix_top20} compares per-code patient-level prevalence for the 20 most common codes in the real training set. MedGAN and HALO most faithfully reproduce the magnitudes of head-of-distribution codes; CorGAN systematically under-predicts even these high-prevalence codes, which is consistent with the near-zero Prevalence $R^2$ at low percentiles in Figure~\ref{fig:longtailed}.

\begin{figure*}[ht]
\centering
\includegraphics[width=\linewidth]{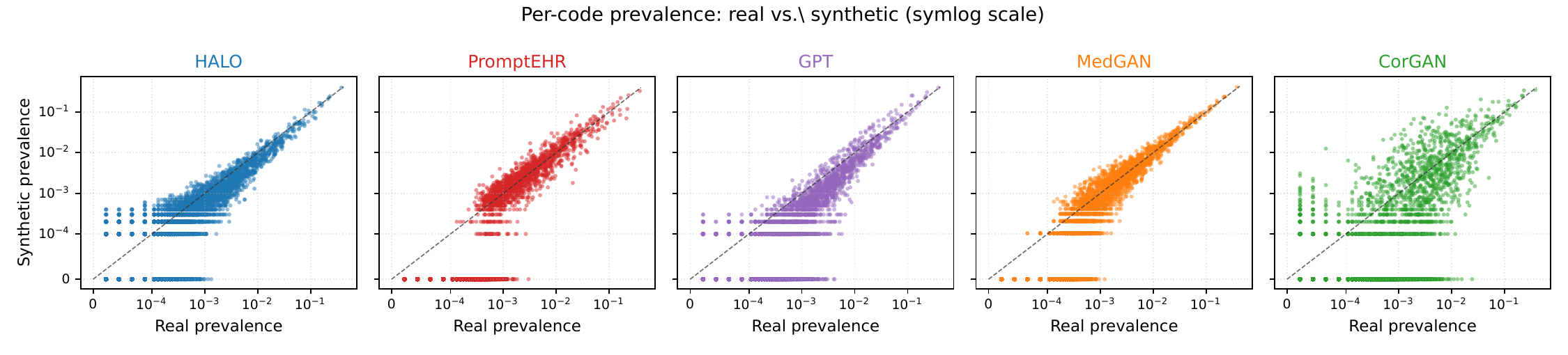}
\caption{Per-code real vs.\ synthetic patient-level prevalence (symlog scale). Each dot is one ICD-9 code. Dashed line is $y = x$.\label{fig:appendix_scatter}}
\end{figure*}

\paragraph{Per-code scatter.} Figure~\ref{fig:appendix_scatter} plots per-code real vs.\ synthetic prevalence on a symlog scale. A generator that perfectly matched the marginals would place every point on the diagonal. HALO and MedGAN cluster tightly along the diagonal for common codes; CorGAN shows a systematic downward bias; PromptEHR and GPT show the characteristic ``fan'' of autoregressive sampling—faithful at the head but under-coverage in the tail.

\begin{figure}[ht]
\centering
\includegraphics[width=\linewidth]{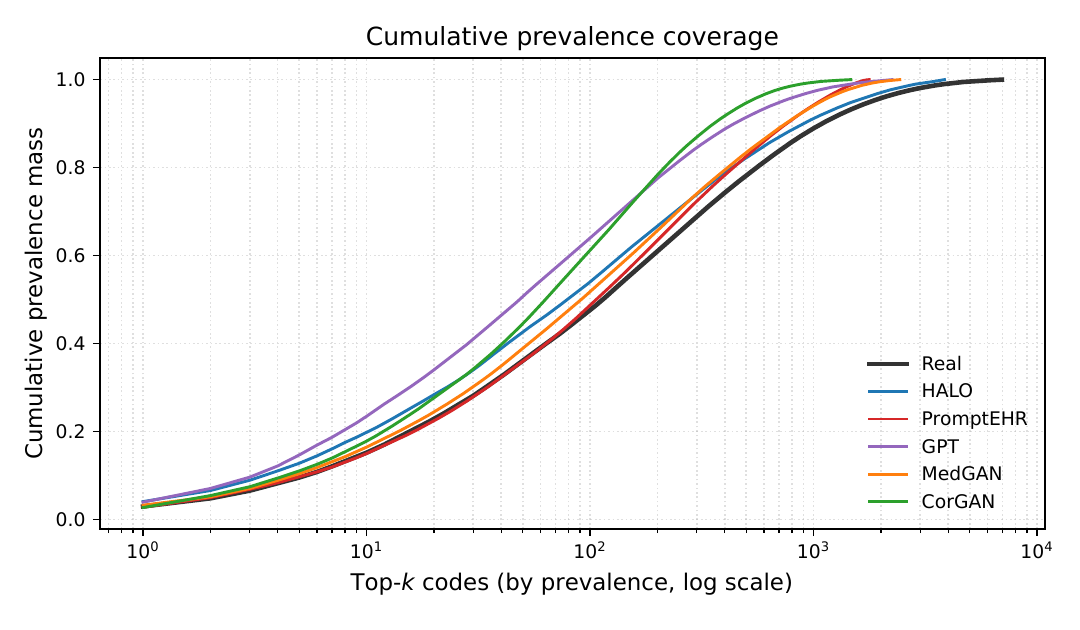}
\caption{Cumulative prevalence mass as a function of the top-$k$ codes. Real MIMIC-III decays smoothly; models that saturate early concentrate too much mass in a narrow head of the distribution.\label{fig:appendix_cumulative}}
\end{figure}

\paragraph{Cumulative prevalence mass.} Figure~\ref{fig:appendix_cumulative} shows the cumulative prevalence mass as a function of the top-$k$ most frequent codes. A steeper curve indicates more mass concentrated in fewer head codes—i.e., a less faithful reproduction of the tail. CorGAN and GPT saturate earliest, consistent with their lower ``unique codes'' counts in Table~\ref{tab:dataset_stats_extra}.

\begin{figure}[ht]
\centering
\includegraphics[width=\linewidth]{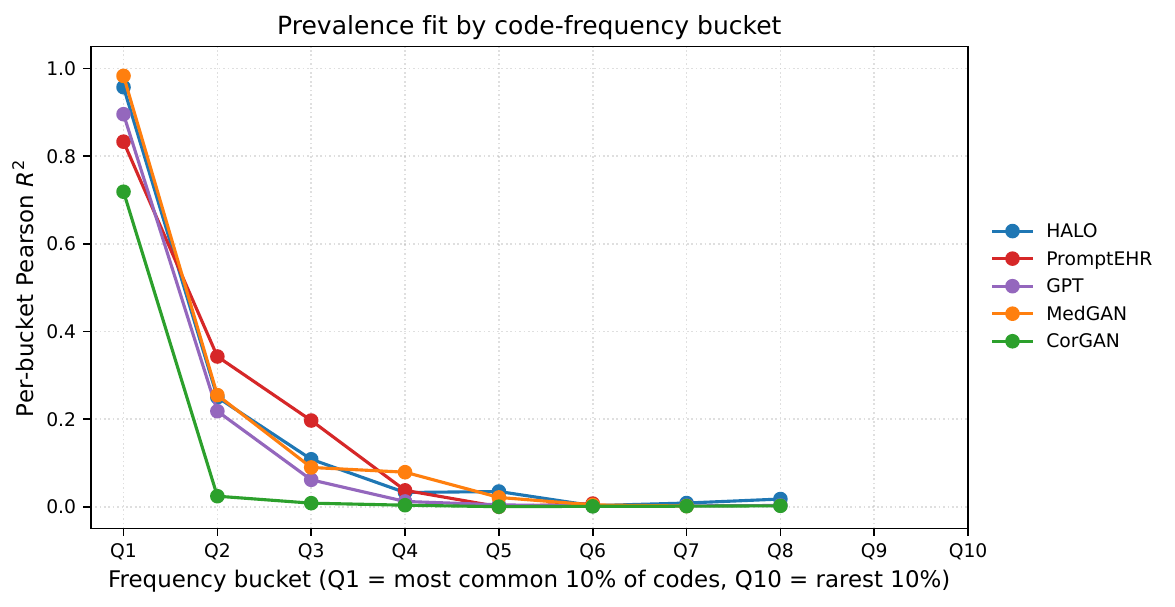}
\caption{Per-bucket Pearson $R^2$ between real and synthetic prevalence, in 10 equal-size frequency buckets (Q1 = most common 10\% of real codes, Q10 = rarest 10\%). The aggregate Prevalence $R^2$ reported in the main text is effectively dominated by the leftmost buckets.\label{fig:appendix_bucket_r2}}
\end{figure}

\paragraph{Per-bucket prevalence fit.} Figure~\ref{fig:appendix_bucket_r2} decomposes the overall Prevalence $R^2$ reported in Table~\ref{tab:synthetic_eval} into ten equal-size frequency buckets, where Q1 contains the 10\% most frequent codes and Q10 contains the rarest 10\%. The aggregate $R^2$ hides a pronounced degradation in the tail: every model's per-bucket $R^2$ falls steeply from Q5 onward, and CorGAN and GPT collapse to near-zero $R^2$ by Q7. This is a finer-grained view of the long-tailed failure flagged in Section~\ref{fig:longtailed}.

\paragraph{Coverage and divergence summary.} Table~\ref{tab:appendix_coverage} consolidates the above into a single scalar view. We report: (i) the number of unique codes each generator ever emits, (ii) the coverage of the real vocabulary, (iii) the Jaccard index between the real and synthetic code sets, (iv) the Kullback--Leibler divergence of the per-code prevalence distributions, (v) the Jensen--Shannon divergence (symmetric, bounded), and (vi) top-$k$ head-overlap for $k \in \{10, 50, 100\}$. HALO has the strongest head-overlap and the lowest KL; CorGAN is consistently the worst match. We emphasize that these are \emph{marginal} distributional metrics and are not a substitute for the discriminator-based fidelity score; they complement it by isolating the vocabulary / coverage axis of model quality.

\begin{table}[ht]
\centering
\caption{Coverage and distributional-divergence metrics between each synthetic sample and the real training set, computed over the per-code patient-level prevalence vector. Coverage is the percentage of real ICD-9 codes that appear at least once in the synthetic sample; KL and JS are over the normalized prevalence distributions.}
\label{tab:appendix_coverage}
\resizebox{\linewidth}{!}{%
\begin{tabular}{lrrrrrrrr}
\toprule
\textbf{Method} & \textbf{Unique} & \textbf{Coverage (\%)} & \textbf{Novel} & \textbf{Jaccard} & \textbf{KL} ($\downarrow$) & \textbf{JS} ($\downarrow$) & \textbf{Top-10} & \textbf{Top-50}  \\
\midrule
HALO         & 3{,}849 & 55.3 & 0 & 0.55 & 0.363 & 0.025 & 8 & 46 \\
PromptEHR    & 1{,}772 & 25.5 & 0 & 0.25 & 0.977 & 0.044 & 6 & 41 \\
GPT Baseline & 2{,}244 & 32.3 & 0 & 0.32 & 1.012 & 0.055 & 8 & 41 \\
MedGAN       & 2{,}429 & 34.9 & 0 & 0.35 & 0.566 & 0.025 & 9 & 46 \\
CorGAN       & 1{,}467 & 21.1 & 0 & 0.21 & 2.707 & 0.150 & 6 & 30 \\
\bottomrule
\end{tabular}%
}
\end{table}

\subsection{Structural ablations: visits and codes-per-visit}
\label{sec:appendix_structure}
Marginal code prevalence is only one axis of fidelity; a high-quality longitudinal generator also has to reproduce how those codes are organized into visits and patients.

\begin{figure}[ht]
\centering
\includegraphics[width=\linewidth]{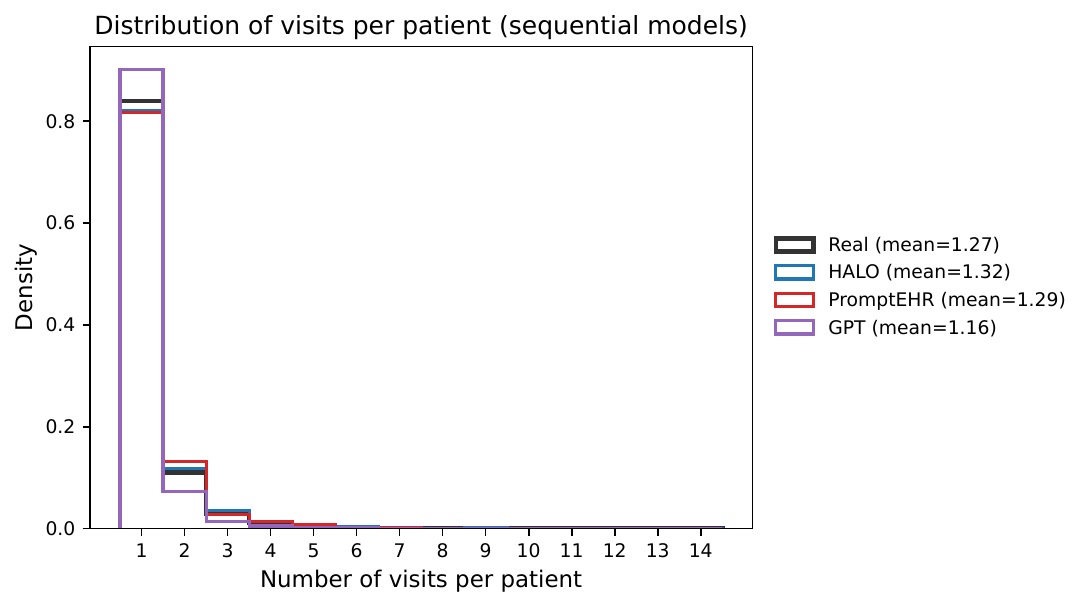}
\caption{Distribution of visits per patient for sequential generators vs.\ real MIMIC-III. Flattened baselines are omitted because they produce one pseudo-visit per patient by construction.\label{fig:appendix_visits}}
\end{figure}

\paragraph{Visits per patient.} Figure~\ref{fig:appendix_visits} shows the distribution of visits per patient for the three sequential generators and the real training data. All three match the real mode of one visit per patient, but GPT slightly under-samples multi-visit patients (mean 1.16 vs.\ 1.27 for real), while HALO over-samples them (1.32). PromptEHR's demographic-conditioned structure sampling (we re-use empirical visit-count and visit-length distributions during generation, see main text) recovers the closest match. Flattened models (MedGAN, CorGAN) cannot participate in this analysis by construction, which we view as an inherent and irreducible limitation of the paradigm.

\begin{figure}[ht]
\centering
\includegraphics[width=\linewidth]{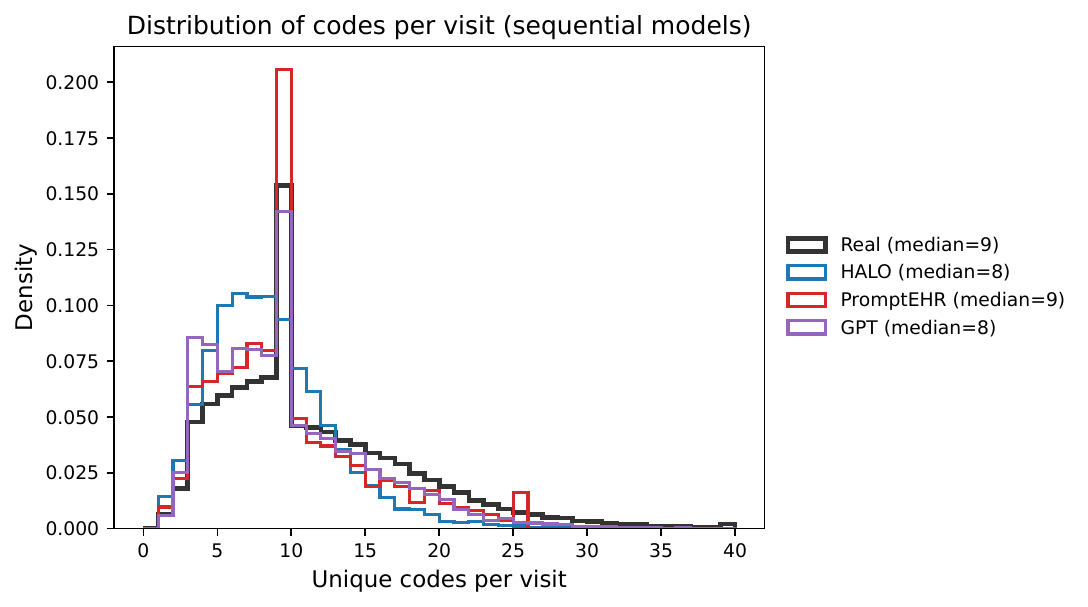}
\caption{Distribution of unique codes per visit for sequential generators. All three models under-cover the upper tail of real visit lengths.\label{fig:appendix_cpv}}
\end{figure}

\paragraph{Codes per visit.} Figure~\ref{fig:appendix_cpv} plots the distribution of unique codes per visit. All three sequential models under-cover longer visits relative to real MIMIC-III: real encounters have a long right tail out to 30+ codes per visit, while HALO/PromptEHR/GPT concentrate their mass below 15 codes. This is a natural consequence of training on teacher-forced next-token prediction: longer visits are penalized more by per-token loss, so the MLE optimum contracts the distribution.

\begin{figure}[ht]
\centering
\includegraphics[width=\linewidth]{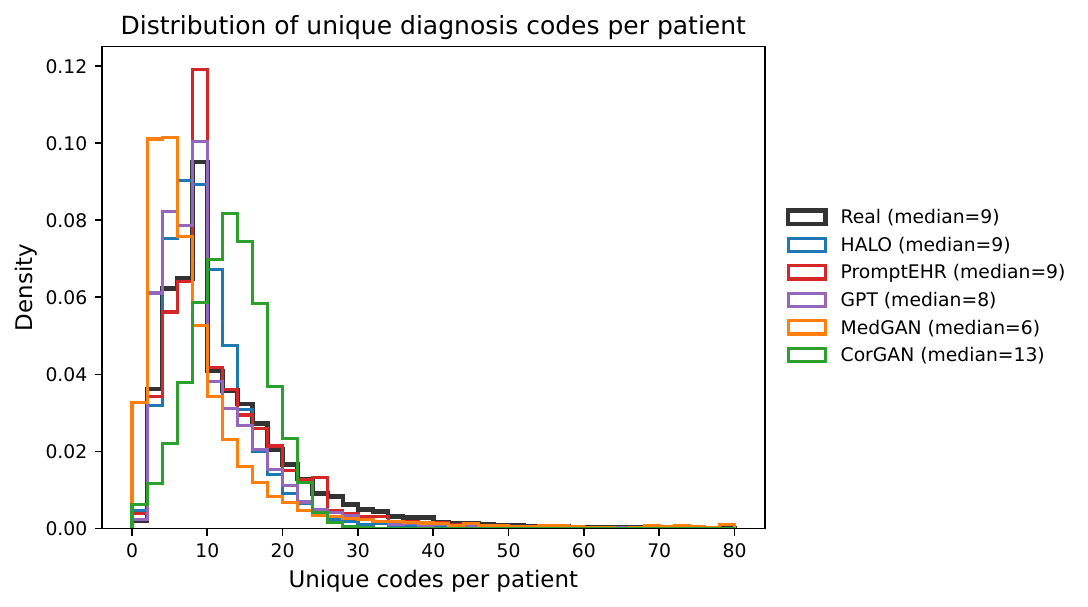}
\caption{Distribution of unique codes per patient. Flattened baselines benefit from pooling all of a patient's codes into a single multi-hot vector, while sequential models must recover this distribution implicitly through visit-level generation.\label{fig:appendix_cpp}}
\end{figure}

\paragraph{Codes per patient.} Figure~\ref{fig:appendix_cpp} is the patient-level analogue of the above: the distribution of unique codes across a patient's full record. PromptEHR is closest to the real median (12 codes) in our sample; HALO and GPT shift left (median 9); flattened baselines---by construction---match real per-patient medians more closely because all of a real patient's codes are pooled into one multi-hot vector, so no sequence-length penalty applies. This highlights a subtle property of the benchmark: flattened models are \emph{aided} by the marginal prevalence metric precisely because they skip the harder sequential-structure task.

\subsection{ICD-9 chapter-level coverage}
\label{sec:appendix_chapters}
\begin{figure*}[ht]
\centering
\includegraphics[width=\linewidth]{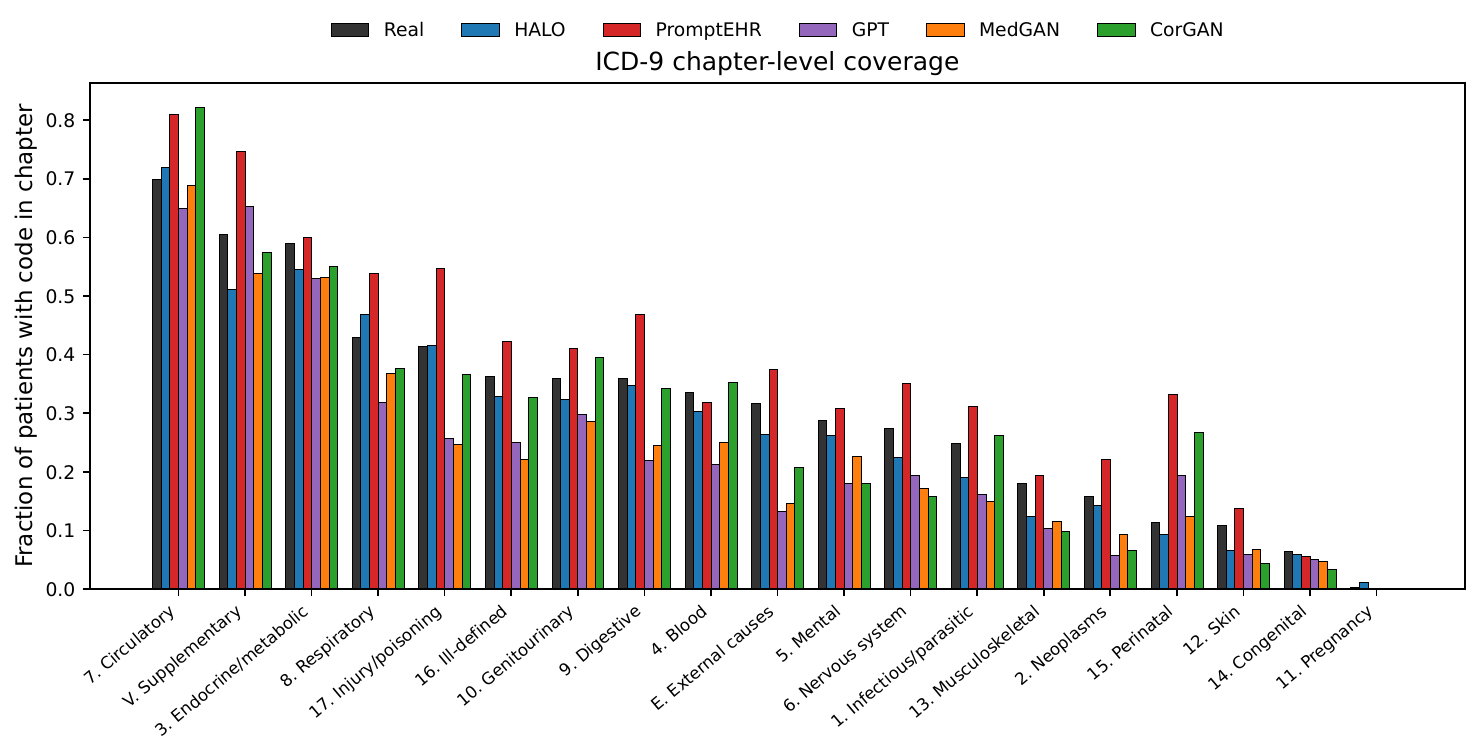}
\caption{Fraction of patients with at least one code in each ICD-9 chapter. This coarse-grained view complements the per-code prevalence analysis by checking for categorical bias.\label{fig:appendix_chapters}}
\end{figure*}

To detect coarse category-level bias, we group codes into ICD-9-CM chapters (Circulatory, Endocrine/Metabolic, Respiratory, Supplementary ``V'' codes, etc.) and plot the fraction of patients whose record contains at least one code in each chapter (Figure~\ref{fig:appendix_chapters}). All models recover the overall shape—Circulatory and Endocrine/Metabolic are the dominant chapters in MIMIC-III—but we observe two consistent biases: (i) GPT and HALO over-emit supplementary ``V'' codes, which are sequence-heavy and appear often at visit boundaries; (ii) CorGAN systematically under-represents chapters beyond the top five, consistent with its poor long-tail behavior. No model visibly over-shoots a clinically implausible chapter.

\subsection{Pairwise code co-occurrence (top-50)}
\label{sec:appendix_cooc}
\begin{figure*}[ht]
\centering
\includegraphics[width=\linewidth]{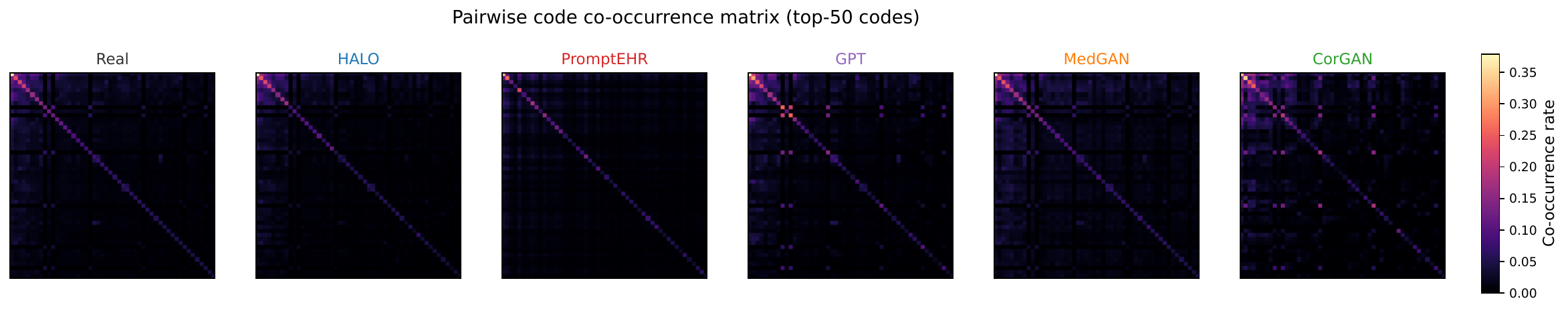}
\caption{Patient-level pairwise code co-occurrence matrix on the 50 most common real ICD-9 codes. Rows/columns are sorted by real prevalence. All heatmaps share the same color scale, set by the real matrix maximum.\label{fig:appendix_cooc}}
\end{figure*}

A generator can match per-code prevalence and still fail to reproduce \emph{co-occurrence} structure (i.e., which codes tend to co-appear in the same patient). Figure~\ref{fig:appendix_cooc} visualizes the patient-level co-occurrence matrix over the 50 most common real codes; Table~\ref{tab:appendix_cooc} reports the cosine similarity and Frobenius difference of each synthetic matrix against the real reference. MedGAN and HALO achieve the highest cosine similarity ($\geq 0.97$), and CorGAN and PromptEHR the lowest ($<0.85$)---indicating that while PromptEHR's aggregate prevalence is reasonable (Table~\ref{tab:synthetic_eval}), its learned second-order structure over the head of the vocabulary is noticeably weaker than HALO's.

\begin{table}[ht]
\centering
\caption{Agreement between real and synthetic pairwise co-occurrence matrices on the top-50 ICD-9 codes. Cosine similarity treats each matrix as a flat vector; Frobenius difference is $\|M_{\text{real}} - M_{\text{syn}}\|_F$.}
\label{tab:appendix_cooc}
\begin{tabular}{lcc}
\toprule
\textbf{Method} & \textbf{Cosine} ($\uparrow$) & \textbf{Frobenius diff} ($\downarrow$) \\
\midrule
HALO         & 0.976 & 0.266 \\
PromptEHR    & 0.836 & 0.660 \\
GPT Baseline & 0.959 & 0.427 \\
MedGAN       & 0.982 & 0.326 \\
CorGAN       & 0.843 & 0.820 \\
\bottomrule
\end{tabular}
\end{table}

\paragraph{Takeaways.} The extended ablations reinforce the conclusion of the main text: (a) flattened models win on marginal prevalence largely because the benchmark metric is also marginal, and they fail structural and co-occurrence fidelity that the discriminator score already flagged; (b) HALO is the most uniformly strong baseline—best long-tail coverage, strongest head co-occurrence, and balanced visit structure—at the cost of its heavier training budget; (c) PromptEHR's aggregate $R^2$ is competitive but is driven almost entirely by head-of-distribution codes (Figure~\ref{fig:appendix_bucket_r2}) rather than broad coverage; (d) the lightweight GPT baseline tracks the sequential frontier closely for common codes but collapses earliest in the long tail, which is consistent with its smaller vocabulary size and relatively cheap training budget. We view these findings as evidence that the single Prevalence $R^2$ / Discriminative Score pair reported in the main table does capture real differences across architectures—but that \emph{no single metric} is sufficient for synthetic EHR benchmarking, and the marginal-plus-structure-plus-co-occurrence triple reported here is the minimum one should inspect.

\end{document}